\documentclass[11pt]{article}

\usepackage[final]{acl}

\usepackage{times}
\usepackage{latexsym}

\usepackage[T1]{fontenc}

\usepackage[utf8]{inputenc}

\usepackage{microtype}

\usepackage{inconsolata}

\usepackage{graphicx}
\usepackage{subcaption}

\newcommand*\circled[1]{\tikz[baseline=(char.base)]{
            \node[shape=circle,draw,inner sep=.6pt] (char) {#1};}}

\usepackage{booktabs}
\usepackage{makecell}  
\usepackage{pifont}    
\usepackage{multirow}
\usepackage{amsfonts}
\usepackage{multicol}
\usepackage{svg}

\usepackage{colortbl}
\newcommand{\cmark}{\ding{51}}
\newcommand{\xmark}{\ding{55}}

\newcommand{\rparagraph}[1]{\vspace{1.4mm}\noindent\textbf{#1.}}

\usepackage[most]{tcolorbox}
\usepackage{tcolorbox}
\newtcolorbox{prompt}{
    colback=black!3,
    colframe=black!40,
    boxrule=0.5pt,
    left=8pt,
    right=8pt,
    top=8pt,
    bottom=8pt,
    arc=2pt,
    breakable,
    enhanced,
    before skip=10pt,
    after skip=10pt
}

\tcbset{
  appendixbox/.style={
    colback=gray!5,
    colframe=gray!50,
    boxrule=0.5pt,
    arc=2pt,
    left=6pt,
    right=6pt,
    top=6pt,
    bottom=6pt
  }
}

\definecolor{darkgreen}{RGB}{0,100,0}
\newcommand{\gap}[1]{\textcolor{darkgreen}{\small(\ensuremath{#1})}}

\newcommand{\std}[1]{{\small \ensuremath{\pm #1}}}

\usepackage{tikz}

\usepackage{titlesec}

\titleformat{\subsubsection}
  {\normalfont\bfseries}
  {\thesubsubsection}
  {0.5em}
  {}

\titlespacing*{\subsubsection}
  {0pt}   
  {0.5ex} 
  {0.5ex} 

\title{Compositional Steering of Large Language Models \\ with Steering Tokens \raisebox{-0.2\height}{\includegraphics[height=1.1em]{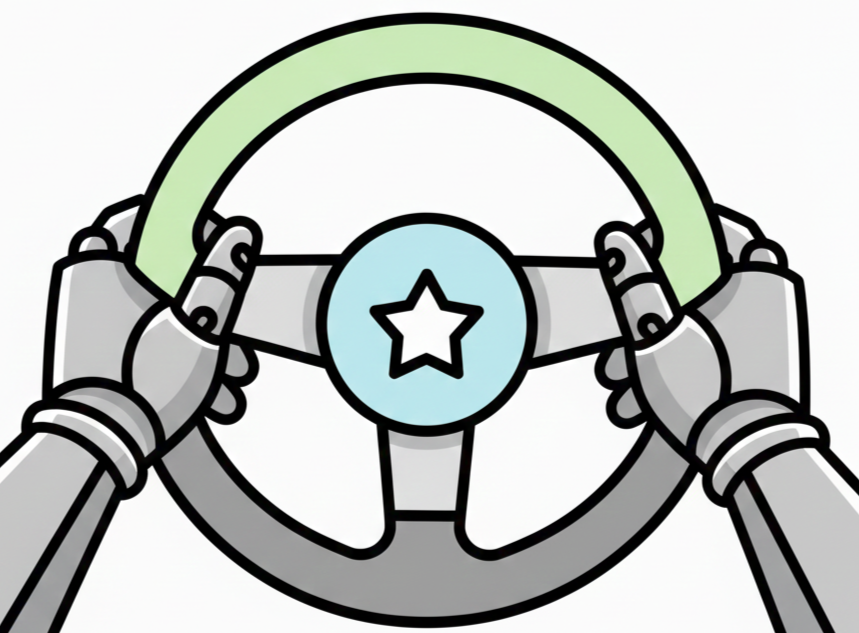}}}

\author{
\textbf{Gorjan Radevski}$^{2}$\thanks{Correspondence to: gorjan.radevski@gmail.com or kiril.gashteovski@neclab.eu},
\textbf{Kiril Gashteovski}$^{1,5}$,
\textbf{Giwon Hong}$^{1,3}$,\\
\textbf{Carolin Lawrence}$^1$,
\textbf{Goran Glavaš}$^4$ 
 \\
$^1$NEC Laboratories Europe, Germany
$^2$Independent \\
$^3$University of Edinburgh, United Kingdom \\
$^4$Center for Artificial Intelligence and Data Science, University of Würzburg, Germany \\
$^5$CAIR, Ss. Cyril and Methodius University of Skopje, North Macedonia \\
}

\begin{document}
\maketitle
\begin{abstract}
Deploying LLMs in real-world applications requires controllable output that satisfies multiple desiderata at the same time. 
While existing work extensively addresses LLM steering for a single behavior, \textit{compositional steering}---i.e., steering LLMs simultaneously towards multiple behaviors---remains an underexplored problem. In this work, we propose \emph{compositional steering tokens} for multi-behavior steering. We first embed individual behaviors, expressed as natural language instructions, into dedicated tokens via self-distillation. Contrary to most prior work, which operates in the activation space, our behavior steers live in the space of input tokens, enabling more effective zero-shot composition. 
We then train a dedicated \textit{composition token} on pairs of behaviors and show that it successfully captures the notion of composition: it generalizes well to \textit{unseen} compositions, including those with unseen behaviors as well as those with an unseen \textit{number} of behaviors. 
Our experiments across different LLM architectures show that steering tokens lead to superior multi-behavior steering of verifiable constraints (e.g., length, format, structure, language) compared to competing approaches (instructions, activation steering, and LoRA merging). Moreover, we show that steering tokens complement natural language instructions, with their combination resulting in further gains. 

\end{abstract}

\section{Introduction}
The ubiquitous use of Large Language Models (LLMs) necessitates that their behavior can be controlled in an ad-hoc manner \cite{khamassi2024strong}. In most real-world domains, due to intrinsic task complexities, the need for \textit{compositional steering}, that is, controlling LLMs for multiple behaviors simultaneously (as opposed to just a single behavior\footnote{We define behavior as any controllable aspect of the LLM's output (e.g., length, language, reasoning style).}), arises naturally \cite{sorensen2024roadmap}. 
%
While fine-tuning LLMs for desired behaviors \cite{tian2023fine,tan-etal-2024-democratizing} is the most straightforward solution to achieve such control, it comes with major limitations. Entailing updates to the LLM's parameters, it (i) is computationally intensive and (ii) jeopardizes the general-purpose utility of the LLM (i.e., there is a risk of negative interference and forgetting of pretrained knowledge); though both can be (somewhat) mitigated via parameter-efficient fine-tuning \cite{hu2022lora}. The need to update the model renders fine-tuning impractical for ad-hoc steering, especially for compositional behavior: $N$ behaviors that can be arbitrarily composed imply $2^N$ (independent) fine-tuning procedures.
%
%
%

The goal of general-purpose instruction-tuning with diverse task instructions \cite{ouyang2022training,zhang2023instruction}, in contrast, is exactly to enable  flexible and generalizable control via natural language prompts. Behavior steering via prompting, however, is very brittle \cite{ngweta-etal-2025-towards,errica2025did}: it yields inconsistent LLM behavior for semantically equivalent prompts.    

Various recent prompt compression approaches \cite{li2025prompt} have been proposed as a middle ground between the inflexibility (and cost) of fine-tuning and the brittleness of prompting, including    
activation steering \cite{stolfo2024improving,cao2024personalized}, gist tokens \cite{mu2023learning,dong2025behavior}, and persona vectors \cite{lim2025persona,pai2025billy}. While these methods efficiently compress prompts and improve LLMs' conformance for individual behaviors, they largely fail to demonstrate effectiveness and generalization in compositional steering.
Put differently, while good solutions for compressing individual behaviors into embeddings are abundant, we still lack an effective representation for the concept of \textit{composition} that would generalize across arbitrary behavior combinations (and \textit{number} of composed behaviors).       


In this work, we address this gap by introducing \emph{compositional steering tokens} for multi-behavior steering (Fig.~\ref{fig:main}). We first compress individual behaviors, formulated as natural language instructions, into specialized tokens (vectors) via self-distillation: unlike in most related work, where behavior representations interact with model internals, our steering tokens reside in the model's input space, which leads to better zero-shot composition. As a central novelty of our work, we then train a dedicated \textit{composition token} to capture the general concept of behavior composition. After training on a fixed set of behavior pairs, we show that our composition token truly encapsulates the notion of composition, exhibiting strong generalization to: (i) \textit{unseen} compositions of \textit{seen} behaviors, (ii) compositions with \textit{unseen} behaviors, and (iii) compositions of \textit{an unseen number} of behaviors.

We carry out an extensive evaluation on \textit{verifiable} behavioral constraints (response length, language, formatting, etc.), enabling strict automatic assessment of compositional generalization at scale. Specifically, we find that:
\circled{1} Steering tokens achieve superior compositional generalization compared to existing methods,
particularly on zero-shot compositions of behavior combinations (\S\ref{sec:steering-vs-baselines}); \circled{2} Compositional gains generalize across diverse model architectures
with hybrid methods providing universal benefits
(\S\ref{sec:steering-different-models});
\circled{3} Both compositional accuracy and robustness scale with model size (up to the 14B models)
(\S\ref{sec:steering-scalling-larger-models});
\circled{4} A learned \texttt{<and>} composition operator is essential for compositional generalization, and orthogonality regularization proves critical for zero-shot fusion (\S\ref{sec:and-token-ablation});
\circled{5} Steering tokens and text instructions exhibit complementary strengths, with hybrid methods consistently achieving superior accuracy-robustness tradeoffs across all behaviors (\S\ref{sec:steering-behavior-breakdown}).
\section{Related Work}

Behavior steering of LLMs is an active area of NLP research \cite{li2025prompt,xie2025survey}. 
We focus on the lines of work we deem most relevant: activation steering, gist tokens, and the few preliminary attempts at compositional steering. Additionally, in Appendix~\ref{app:soft-prompt-tuning} we discuss how the proposed steering tokens are connected to soft prompt tuning.

\rparagraph{Activation Steering} These methods embed the behavior into vectors that are combined with models' activations (i.e., token representations) via simple arithmetic operations. Most activation steering methods are \textit{contrastive}: steering vectors are obtained by comparing activations under exhibited behavior with some type of baseline activations, either in the original activation space \cite{rimsky2024steering,stolfo2024improving,im2025unified,sterz-etal-2025-recover,wu2025improved} or a derived one (e.g., PCA or sparse decomposition) \cite{siddique2025shifting,bayat2025steering}. The most relevant for our work is that of \newcite{stolfo2024improving}, who obtain the activation steers by contrasting LLM's activations for two different prompts: \textit{with} vs. \textit{without} the behavior instruction. We adopt the same two-prompt approach, but distill the instruction into the steering token (appended to the prompt without the behavior instruction) by contrasting the distributions of the two outputs (see \S\ref{sec:method} and Figure \ref{fig:main}). Critically, while the steering vectors of \citet{stolfo2024improving} improve instruction following \textit{on top of} prompt steering (i.e., behavior instruction in the prompt), \textit{alone} they are dramatically worse than prompt-based steering. In stark contrast, our steering tokens for individual behaviors are, alone, as effective as instruction steering (see Table~\ref{tab:single-token-performance} in Appendix~\ref{app:extra-experiments}).          

\rparagraph{Gist Tokens} Concerned with test-time compute efficiency, this line of work compresses complex behaviors (i.e., behaviors described by long instructions) into a few input tokens, via various objectives \cite{mu2023learning,han2024word,petrov2025long}. \newcite{mu2023learning}, for example, train them by modifying the self-attention patterns (gist tokens attend over the behavior description, and answer tokens over gist tokens), whereas the concurrent work of \newcite{dong2025behavior} adopts a self-distillation approach, which is conceptually similar to ours, albeit more complex (see \S\ref{sec:method}). Gist tokens have been used in various use-cases, including for better selection of in-context examples \cite{gupta2023gistscore}, conditioned decoding \cite{li2024say}, hierarchical compression for longer contexts (by gisting the gist tokens) \cite{petrov2025long,tarasov2025sentence}, and in task-specific gisting \cite{phang2024investigating,jiang2024hierarchical}. Gist approaches, however, tackle only single-instruction steering, whereas we are primarily interested in behavior compositions. 

\rparagraph{Compositional Steering} While compositional steering remains much less studied than steering for individual behaviors, a few efforts in multi-behavior steering do exist.   
Most attempts merely interpolate between individual steering vectors to obtain compositional behavior \cite{han2024word,scalena2024multi,cao2024personalized}.  
Directly composing independently trained modules on top of the same LLM parameters or activations, however, is known to be destructive; this favors sparse changes to the underlying LLM \cite{ansell2024scaling,van2024extending}. \newcite{stolfo2024improving} insert different steers in different LLM layers. \newcite{nguyen2025multi} sparsify individual steering vectors and force their orthogonality, thereby mitigating the negative interference in interpolation-based composition. The orthogonality constraint, however, demands that all behaviors are known in advance; addition of a new behavior requires recomputation of all steering vectors. In contrast, our approach yields compositional generalization: we train a dedicated composition token that learns the very function of composing behaviors and can thus be applied to behaviors unseen in its training.   

Critically, these efforts lack rigorous evaluation of multi-behavior steering. Most focus on single-behavior steering and address composition merely as a side experiment, providing only anecdotal evidence that some behavior compositions work \cite{stolfo2024improving,cao2024personalized,han2024word}. \newcite{scalena2024multi} provide a quantitative evaluation but no baseline performance, whereas \newcite{nguyen2025multi} compare only against in-context learning and fine-tuning approaches (SFT and DPO); both fail to include the simplest, yet very competitive baseline (cf. \S\ref{sec:results}): compositional steering in the prompt (i.e., via instruction). In contrast, we perform a rigorous and extensive empirical evaluation of multi-behavior steering with automatically verifiable behaviors (\S\ref{sec:experimental-setup}).  

\section{Compositional Steering with Tokens}\label{sec:method}
\begin{figure*}[t]
  \centering
  \includegraphics[width=\linewidth]{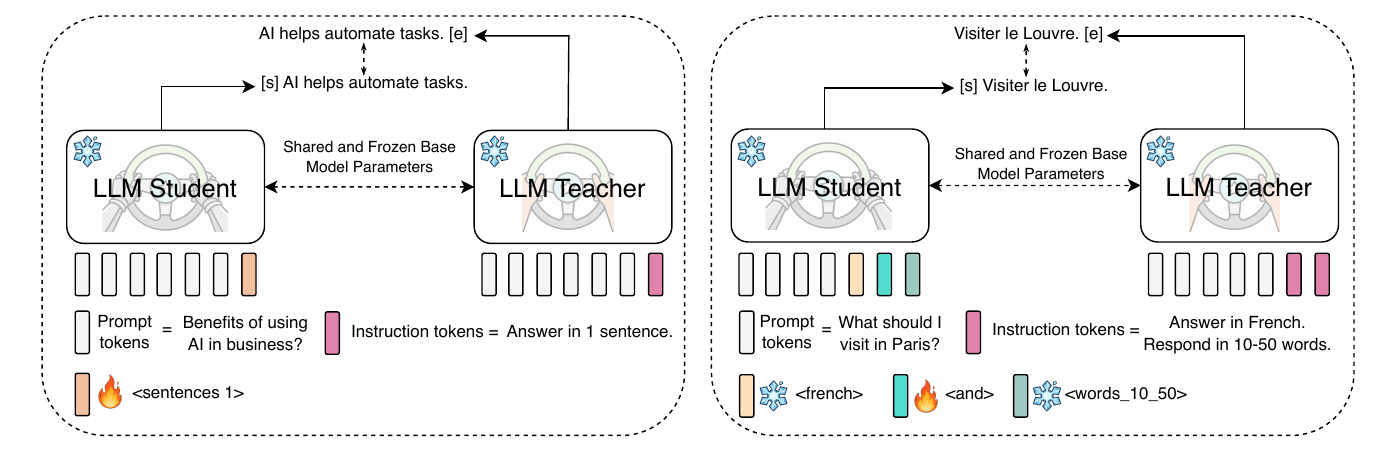}
  \caption{Illustration of our compositional self-distillation. \textbf{Left:} \protect\circled{1} Training individual behavior steering tokens; \textbf{Right:} \protect\circled{2} Training the composition token \texttt{<and>}. The LLM remains fully frozen (including its subword embeddings). We \textit{self-distill} from the instruction-prompted LLM to train the respective steering token:  \textit{single} behavior tokens in \protect\circled{1}, and the \texttt{<and>} token in \protect\circled{2}.}
  \label{fig:main}
  \vspace{-1em}
\end{figure*}


We tackle multi-behavior steering, where LLMs need to generate output that simultaneously conforms to multiple constraints. Formally, given a set of behaviors  $\mathcal{B} = \{b_1, b_2, \ldots, b_k\}$ (e.g., $b_i=$\textit{``Answer in one sentence.''}; $b_j=$\textit{``Answer in French.''}) and an input prompt $x$ (e.g., $x=$ \textit{``Why penguins can't fly?''}), expect the LLM to produce a high-quality answer $y$ for the prompt $x$, while satisfying all constraints from $\mathcal{B}$. 
Standard instruction-based steering would simply concatenate behavior descriptions (e.g., ``Answer in French. Use one sentence.''), but is known to be brittle \cite{jaroslawicz2025many}. We propose \textit{compositional steering tokens}: input embeddings for compositional steering that do not change models' internals.  

\rparagraph{Steering Tokens: Input-Level Control}
Our steering tokens operate in the input embedding space; we keep the base LM frozen. For each behavior $b \in \mathcal{B}$, we introduce a trainable \textit{steering token} \texttt{<b>}, that is, an embedding $\mathbf{e}_b \in \mathbb{R}^{d}$, with $d$ as the model's hidden size. Additionally, we add a trainable \textit{composition token} \texttt{<and>}, i.e., a vector $\mathbf{e}_{\text{<and>}} \in \mathbb{R}^{d}$ in which we embed the notion of composition itself.   
For a prompt $x$ constrained with two behaviors $\{b_i, b_j\}$, we feed the following concatenation of tokens to the LLM: $[\mathbf{E}_x, \mathbf{e}_{b_i}, \mathbf{e}_{\text{<and>}}, \mathbf{e}_{b_j}]$, where $\mathbf{E}_x$ denotes the sequence of (frozen) embeddings of $x$'s tokens. This design 
(i) prevents model collapse, as keeps all LLMs' parameters frozen, (ii) facilitates composition, as we combine behaviors through learned interactions in the input space; and (iii) is computationally efficient: we only need to learn $|\mathcal{B}|+1$ $d$-dimensional vectors: one for each behavior and one for the composition token.

\rparagraph{Compositional Self-Distillation}
A key challenge in multi-behavior control is enabling \textit{compositional generalization}: the ability to combine behaviors at inference time without training on all possible behavior combinations. To this end, our proposal is the compositional operator (i.e., token) \texttt{<and>}, meant to mediate
interactions between behavior embeddings. We thus train in two stages: \circled{1} we independently train individual behavior tokens \texttt{<b>}, keeping the LLM frozen (left side of Fig.~\ref{fig:main}), and then \circled{2} train the composition token \texttt{<and>} (right side of Fig.~\ref{fig:main}) on different two-behavior combinations $b_i$ and $b_j$ (unless specified otherwise), keeping both the LLM and the behavior embeddings \texttt{<b\textsubscript i>} and \texttt{<b\textsubscript j>} \textit{frozen}. Keeping the behavior tokens frozen in the second step is key for \textit{compositional generalization} as it ensures that \texttt{<and>} learns the behavior-independent concept of \textit{composition}, instead of just modifying individual behavior representations.

We train both the individual behavior tokens \texttt{<b>} and the composition token \texttt{<and>} via self-distillation. For individual behaviors $b$, the input to the teacher is the prompt $x$ with a behavior's natural language instruction $I_b$, whereas the student receives $x$ appended with a steering token \texttt{<b>}. The teacher and student are the very same frozen LLM, and they both need to output the same answer $y$. Our distillation objective minimizes the KL-divergence between the teacher's and student's output distributions over answer tokens: 
%
\begin{equation*}
\mathcal{L}_{\text{dist}} \hspace{-0.1em}=\hspace{-0.1em} \text{KL}\hspace{-0.1em}\left(P_{\text{teacher}}(y \hspace{-0.2em}\mid \hspace{-0.2em} x, I_b) \| P_{\text{student}}(y \hspace{-0.2em}\mid \hspace{-0.2em} x, \texttt{\small{<b>}})\right)  
\end{equation*}


\noindent where both distributions are temperature-scaled with $T$: $P(y) = \text{softmax}\left({\text{logits}(y)}/{T}\right)$. We scale the final loss by $T^2$ to account for temperature normalization \cite{radevski2023multimodal}. In our experiments, we use high temperature ($T = 10.0$) to encourage the student to match the full probability distribution rather than just the mode.
To prevent overfitting of behavior tokens to instruction formulations, we define 10 different instruction paraphrases for each behavior (e.g., \textit{``\{Answer, Respond, Reply\} in Spanish''}) and randomly sample one per training example.

When training the composition token \texttt{<and>}, the teacher receives $x$ concatenated with instructions of two behaviors, $x \oplus I_{b_i} \oplus I_{b_j}$ (e.g., \textit{``Why can't penguins fly? Answer in Spanish. Use 10 to 50 words.''}) and the student receives $x \oplus \texttt{<b\textsubscript i>} \texttt{<and>} \texttt{<b\textsubscript j>}$ (e.g., \textit{``Why can't penguins fly?}\,\texttt{<ES><and><words\_10\_50>}\textit{''}.

\rparagraph{Initialization and Orthogonality} 
We initialize embeddings of behavior tokens \textit{``semantically''}: as the mean of the LLM's (frozen) embeddings of tokens in the behavior's instruction. 
We use zero initialization for the embedding of the composition token (we ablate this choice in \S\ref{sec:and-token-ablation}): $\mathbf{e}_{\text{<and>}} = \mathbf{0}$, to avoid biasing it toward any one behavior and allowing it to learn ``to compose'' purely from the training data. 

To prevent the \texttt{<and>} token from collapsing into representations similar to existing behavior tokens, we introduce \textit{orthogonality regularization} in composition training (we ablate its impact in \S\ref{sec:and-token-ablation}): it forces the trainable \texttt{<and>} embedding to remain orthogonal to \textit{all} frozen behavior embeddings:
\begin{equation*}
    \mathcal{L}_{\text{orth}} = \sum_{b \in \mathcal{B}_{\text{seen}}} \left({\mathbf{e}_{\text{<and>}} \cdot \mathbf{e}_b} / {(\|\mathbf{e}_{\text{<and>}}\|\cdot\|\mathbf{e}_b\|)}\right)^2
\end{equation*}
\noindent where $\mathcal{B}_{\text{seen}}$ is the set of all (frozen) behavior tokens \textit{seen} in composition training. The final loss for the \texttt{<and>} token training is then: $\mathcal{L} = \mathcal{L}_{\text{dist}} + \lambda \cdot \mathcal{L}_{\text{orth}}$. We initially set $\lambda = 0.5$ and found it to work well. 

\section{Experimental Setup}
\label{sec:experimental-setup}

Unlike work on steering for subjective behaviors such as sycophancy or myopic reward \cite{nguyen2025multi, bayat2025steering, scalena2024multi, pai2025billy}, we follow \cite{stolfo2024improving} and focus on \textit{verifiable} constraints where satisfaction can be automatically assessed, allowing for a large-scale compositional evaluation while avoiding both human evaluation and costly and error-prone use of LLMs as judges \cite{marioriyad2025silent}. 



\rparagraph{Models} We experiment with seven instruction-tuned LLMs from four model families: Qwen3-4B, Qwen3-8B, Qwen3-14B, Llama-3.2-3B, Llama-3.2-8B, SmolLM3-3B, OLMo-7B, and Gemma3. We use Qwen-8B as the primary model for comparing steering tokens with baselines (\S\ref{sec:steering-vs-baselines}) and the Qwen family models  
for the scaling analysis (\S\ref{sec:steering-scalling-larger-models}). For efficiency, we use Qwen3-4B in ablations (\S\ref{sec:and-token-ablation}).

\rparagraph{Behaviors} We experiment with 15 behaviors from four categories: \textit{languages}: Spanish, French, Italian, Portuguese, and German; \textit{length}: 10-50, 50-70, 70-90, 90-120 words; \textit{formatting}: lowercase, uppercase, title case; and \textit{structure}: 1-5 sentences. 
To test compositional generalization, we partition properties into \textit{seen} (11 used during \texttt{<and>} token training) and \textit{unseen} (4 held out: German, title case, 70-90 words, 3 sentences). This design ensures that each category has held-out properties to validate zero-shot composition. We provide instructions for all behaviors in the Appendix~\ref{app:instruction-variants}.

\rparagraph{Compositions} We evaluate two types of behavior combinations: (i) \textbf{Seen compositions}: both behaviors are from the seen set, i.e., their combination was part of the \texttt{<and>} token training (e.g., \texttt{<ES> <and> <words\_10\_50>}); here we test if the composition operator successfully learned to compose seen behaviors; (ii) \textbf{Unseen compositions}: one or both behaviors are from the \textit{unseen} behavior set (e.g., \texttt{<DE> <and> <words\_10\_50>}, \texttt{<ES> <and> <title\_case>} or \texttt{<DE> <and> <title\_case>}: this setup tests the generalization ability of our composition token operator.\footnote{The seen/unseen split is defined relative to the \texttt{<and>} token's training data; instruction steering inherits it for comparability. Any gap there reflects genuine difficulty variation: e.g., German paired with \texttt{title\_case} conflicts with German's capitalization of all nouns.}

In both cases, we test on 2-behavior and 3-behavior compositions. Crucially, 3-behavior compositions are \textit{all} \textit{unseen}, since we train the \texttt{<and>} only on 2-behavior combinations, providing a different test of generalization---to compositions with different \textit{number} of behaviors. For the Qwen family, we contrast our default training of the composition token \texttt{<and>} exclusively on 2-behavior compositions (\textbf{2-token-only}), against training on both 2-behavior and 3-behavior compositions (\textbf{2+3-token}): this way, we evaluate whether explicit training on 3-behavior compositions improves generalization or causes overfitting.


\rparagraph{Data} We source prompts from the Smoltalk dataset\footnote{The dataset is released under Apache 2.0 License.} \cite{allal2025smollm2} which contains instruction-labeled dialogs. We use  Qwen3-30B-A3B-Instruct to separate the core question for each prompt from constraints (e.g., \textit{``respond in 5 sentences,''}).
For each of our 15 behaviors, we randomly sample 50k prompts and generate answers with Qwen3-30B-A3B-Instruct (for each example, we randomly sample from 10 behavior paraphrases). For the \texttt{<and>} token training, we generate responses for all cross-category 2-behavior combinations (e.g., \textit{language} + \textit{length}, \textit{language} + \textit{format}). 
We exclude all unseen behaviors (German, title case, 70-90 words, 3 sentences) from the training data creation and save them for evaluation of \textit{unseen} compositions (see App.~\ref{app:implementation-details} for further details).
For testing, we use 1,000 held-out prompts for each 2- and 3-behavior combination, resulting in >1M evaluations per model across all compositions and orderings of behavior tokens.

\rparagraph{Metrics} We measure the following: (1) \textit{Mean accuracy}, as the percentage of LLM generations that satisfy all $k \in {2, 3}$ behaviors, averaged across all $k!$ token orders: this way, we remove any potential model bias w.r.t. the order of behavior tokens;   
(2) \textit{Order variance} is the largest absolute difference in accuracy across any two token orders of a composition, averaged across all compositions. Lower values indicate more robust, order-invariant behavior; 
and (3) \textit{Response quality}: for \textit{accurate} responses (i.e., satisfying all behaviors) constraints, we evaluate semantic correctness and coherence of the answers on a Likert (1-5) scale, using an LLM judge (Qwen3-30B-A3B-Instruct). We measure whether steering accuracy comes at the expense of degraded content (see details in Appendix.~\ref{app:llm-as-judge}).

\begin{table*}[t]
    \centering
    \resizebox{1.0\textwidth}{!}{
    \begin{tabular}{l cccc c cccc}
        \toprule
        & \multicolumn{4}{c}{\textbf{2-Behavior Composition}} & & \multicolumn{4}{c}{\textbf{3-Behavior Composition}} \\
        \cmidrule(lr){2-5} \cmidrule(lr){7-10}
        \textbf{Method} & \textbf{Seen $\uparrow$} & \textbf{Unseen $\uparrow$} & \textbf{Ord. Var. $\downarrow$} & \textbf{Resp. Qual. $\uparrow$} & & \textbf{Seen $\uparrow$} & \textbf{Unseen $\uparrow$} & \textbf{Ord. Var. $\downarrow$} & \textbf{Resp. Qual. $\uparrow$} \\
        \midrule
        \textit{\textbf{Baselines}} \\
        CAA \cite{rimsky2024steering} & 1.6 \std{0.1} & 0.5 \std{0.1} & -- & 1.1 & & 0.6 \std{0.1} & 0.1 \std{0.1} & -- & 1.1 \\
        LM-Steer \cite{han2024word} & 18.1 \std{0.1} & 13.4 \std{0.3} & -- & 1.3 & & 2.2 \std{0.1} & 2.1 \std{0.2} & -- & 1.2 \\
        LoRA DARE \cite{yu2024language} & 81.5 \std{0.1} & 44.8 \std{0.2} & -- & 4.7 & & 58.4 \std{0.1} & 17.6 \std{0.1} & -- & 4.6 \\
        Instruction Steering \cite{stolfo2024improving} & 90.7 \std{0.1} & 71.8 \std{0.2} & 7.8 & 4.9 & & 83.7 \std{0.1} & 54.0 \std{0.1} & 18.1 & 4.9 \\
        \midrule
        \textit{\textbf{Steering Tokens}} \\
        Concatenation & 81.3 \std{0.1} & 62.1 \std{0.2} & 25.3 & 4.8 & & 59.6 \std{0.1} & 33.2 \std{0.1} & 55.8 & 4.8 \\
        Composition (native ``and'' token) & 82.9 \std{0.1} & 66.4 \std{0.2} & 12.9 & 4.8 & & 68.8 \std{0.1} & 47.1 \std{0.2} & 40.7 & 4.8 \\
        \rowcolor{green!5}Composition (\texttt{<and>}) & 90.9 \std{0.1} & \textbf{76.9 \std{0.2}} & 5.3 & 4.9 & & 83.1 \std{0.1} & 59.5 \std{0.1} & 25.5 & 4.9 \\
        \rowcolor{green!5}Hybrid: Composition (\texttt{<and>}) + Instr. & \textbf{92.2 \std{0.1}} & 76.3 \std{0.2} & \textbf{4.4} & 4.9 & & \textbf{87.9 \std{0.1}} & \textbf{62.9 \std{0.1}} & \textbf{15.2} & 4.9 \\
        \bottomrule
    \end{tabular}
    }
    \caption{Results for compositional steering with Qwen3-8B for \textit{seen} and \textit{unseen} 2- and 3-behavior combinations. The learned \texttt{<and>} composition operator outperforms text instructions on unseen compositions; Combining our token steering with instructions (Hybrid) yields the best performance; Concatenation without explicit composition learning fails on complex compositions; LoRA DARE \cite{yu2024language} shows poor compositional generalization, whereas LM-Steer \cite{han2024word} and and CAA \cite{rimsky2024steering} fails to compose altogether.}
    \label{tab:steering-vs-baselines}
    \vspace{0.5em}
\end{table*}

\rparagraph{Baselines and Variants} We compare our steering tokens against four baselines that instantiate different types of steering.
\textbf{(1)} In practice, \textit{instruction steering} is the default paradigm for multi-behavior steering of LLMs; despite that, existing work on compositional steering \cite{cao2024personalized,han2024word,nguyen2025multi}, with the exception of \cite{stolfo2024improving}, fails to evaluate this competitive baseline. We simply append the behavior instructions $I_b$ to the prompt (e.g., \textit{``Answer in German. Use 70-90 words. Apply title case.''}). To ensure fair comparison, we randomly sample from 10 behavior paraphrases.
\textbf{(2)} Standard activation steering as per \citet{rimsky2024steering}\footnote{\newcite{stolfo2024improving} empirically shows that the steering vectors by \citet{rimsky2024steering}, when applied in a standalone fashion clearly and substantially trail instruction-based steering for \textit{verifiable} behaviors.}
\textbf{(3)} Fine-tuning for behavior alignment is an expensive but effective steering approach. We first train behavior-specific low-rank adapters with the same self-distillation objective, then merge individual adapters using an interference-reducing approach of \newcite{yu2024language} (\textit{LoRA DARE}); this is a meaningful parameter-space alternative to our input-space steering.
\textbf{(4)} LM-Steer \cite{han2024word} is an established approach with steering vectors obtained as linear projections of LLMs' word embeddings: as such, it is essentially a type of token-based steering; moreover, the authors claim (but do not quantify) LM-Steer's compositional steering abilities.
Check Appendix~\ref{app:baselines} for implementation details of the baselines.

We evaluate four variants of the steering token approach: \textbf{(1)} a simple \textit{concatenation} of behavior tokens (i.e., no composition token \textit{<and>}), shedding light on the necessity of explicit composition learning;
\textbf{(2)} using the native ``and'' LLM token embedding as a composition token, in order to understand whether the trained \textit{<and>} operator encodes something qualitatively different from the word ``and''; \textbf{(3)} the default \textit{<and>} composition; and \textbf{(4)} we combine our token steering with instructions (\textit{Hybrid}), testing the complementarity of learned vectors and natural language guidance.
\section{Results and Analyses}
\label{sec:results}

\begin{table}[t]
    \centering
    \resizebox{1.0\columnwidth}{!}{
    \begin{tabular}{ll cc}
        \toprule
        \textbf{Model} & \textbf{Method} & \textbf{Unseen (2|3) $\uparrow$} & \textbf{Order Var. (2|3) $\downarrow$} \\
        \midrule
        
        \multirow{3}{*}{Qwen3-4B} 
        & Instruction & 68.9 | 55.6 & 5.3 | \textbf{15.1} \\
        & Steering & 69.1 | 60.7 & 6.2 | 22.7 \\
        & \cellcolor{green!5}Hybrid & \cellcolor{green!5}\textbf{69.2 \gap{+0.3}} | 58.0 \gap{+2.4} & \cellcolor{green!5}\textbf{5.3 \gap{+0.0}} | 18.6 \gap{+3.5} \\
        \midrule
        
        \multirow{3}{*}{Qwen3-8B} 
        & Instruction & 71.8 | 54.0 & 5.1 | 15.1 \\
        & Steering & \textbf{76.9} | 59.5 & \textbf{4.1} | 21.2 \\
        & \cellcolor{green!5}Hybrid & 76.3 \gap{+4.5} | \cellcolor{green!5}\textbf{62.9 \gap{+8.9}} & 4.4 \gap{-0.7} | \cellcolor{green!5}\textbf{12.4 \gap{-2.7}} \\
        \midrule
        
        \multirow{3}{*}{Llama3-3B} 
        & Instruction & 66.7 | 33.8 & 3.1 | \textbf{8.7} \\
        & Steering & 69.3 | 33.9 & 3.9 | 18.7 \\
        & \cellcolor{green!5}Hybrid & \cellcolor{green!5}\textbf{74.9 \gap{+8.2}} | \cellcolor{green!5}\textbf{43.4 \gap{+9.6}} & \cellcolor{green!5}\textbf{2.8 \gap{-0.3}} | 10.6 \gap{+1.9} \\
        \midrule
        
        \multirow{3}{*}{Llama3-8B} 
        & Instruction & 67.8 | 40.2 & 3.6 | 13.6 \\
        & Steering & 67.0 | 39.5 & 5.9 | 19.2 \\
        & \cellcolor{green!5}Hybrid & \textbf{76.3 \gap{+8.5}} | \cellcolor{green!5}\textbf{52.9 \gap{+12.7}} & \textbf{3.2 \gap{-0.4}} | \cellcolor{green!5}\textbf{11.0 \gap{-2.6}} \\
        \midrule
        
        \multirow{3}{*}{Smol3-3B} 
        & Instruction & 53.2 | 32.5 & \textbf{5.1} | \textbf{14.5} \\
        & Steering & 53.2 | 35.5 & 11.8 | 35.9 \\
        & \cellcolor{green!5}Hybrid & \cellcolor{green!5}\textbf{53.5 \gap{+0.3}} | \cellcolor{green!5}\textbf{37.2 \gap{+4.7}} & \cellcolor{green!5}7.3 \gap{+2.2} | 17.1 \gap{+2.6} \\
        \midrule
        
        \multirow{3}{*}{Olmo-7B} 
        & Instruction & 56.8 | 30.9 & \textbf{2.9} | 7.3 \\
        & Steering & 56.9 | 28.4 & 3.6 | 12.3 \\
        & \cellcolor{green!5}Hybrid & \cellcolor{green!5}\textbf{60.9 \gap{+4.1}} | \textbf{37.5 \gap{+6.6}} & \cellcolor{green!5}3.7 \gap{+0.8} | \textbf{6.4 \gap{-0.9}} \\
        \midrule

        \multirow{3}{*}{Gemma3-4B} 
        & Instruction & 39.9 | 21.1 & \textbf{3.4} | \textbf{6.1} \\
        & Steering & 37.4 | 18.8 & 4.9 | 14.0 \\
        & \cellcolor{green!5}Hybrid & \cellcolor{green!5}\textbf{49.5 \gap{+9.6}} | \textbf{26.7 \gap{+5.6}} & \cellcolor{green!5}6.5 \gap{+3.1} | 8.2 \gap{+2.2} \\
        
        \bottomrule
    \end{tabular}
    }
    \caption{Cross-architecture generalization of steering tokens, text instructions, and hybrid methods across seven models spanning four architectural families. Hybrid methods consistently achieve the best performance across all architectures, with strong benefits on weaker compositional models (Llama, OLMo). The compositional advantage holds universally but shows architecture-dependent magnitudes. Numbers in brackets represent improvement on top ``Instruction'' baseline. See App.~\ref{app:extra-experiments} for performance on seen categories.}
    \label{tab:steering-different-models}
    \vspace{0.5em}
\end{table}

\begin{table*}[t]
    \centering
    \resizebox{1.0\textwidth}{!}{
    \begin{tabular}{ll cccc c cccc}
        \toprule
        & & \multicolumn{4}{c}{\textbf{2-Behavior Composition}} & & \multicolumn{4}{c}{\textbf{3-Behavior Composition}} \\
        \cmidrule(lr){3-6} \cmidrule(lr){8-11}
        \textbf{Model} & \textbf{Method} & \textbf{Seen $\uparrow$} & \textbf{Unseen $\uparrow$} & \textbf{Ord. Var. $\downarrow$} & \textbf{Resp. Qual. $\uparrow$} & & \textbf{Seen $\uparrow$} & \textbf{Unseen $\uparrow$} & \textbf{Ord. Var. $\downarrow$} & \textbf{Resp. Qual. $\uparrow$} \\
        \midrule
        
        \multirow{3}{*}{Qwen3-4B}
        & Instruction & 93.3 & 68.9 & \textbf{5.3 \gap{+0.0}} & 4.8 & & 88.2 & 55.6 & \textbf{15.1 \gap{+0.0}} & 4.8 \\
        & Steering (2-only) & 93.7 & 69.1 & 6.2 & 4.8 & & 89.3 & \textbf{60.7 \gap{+5.1}} & 22.7 & 4.8 \\
        & \cellcolor{green!5}Hybrid & \cellcolor{green!5}\textbf{93.7 \gap{+0.4}} & \cellcolor{green!5}\textbf{69.2 \gap{+0.3}} & \cellcolor{green!5}\textbf{5.3 \gap{+0.0}} & \cellcolor{green!5}4.8 &\cellcolor{green!5} & \cellcolor{green!5}\textbf{90.7 \gap{+2.5}} & \cellcolor{green!5}58.0 & \cellcolor{green!5}18.6 & \cellcolor{green!5}4.8 \\
        \midrule
        
        \multirow{4}{*}{Qwen3-8B}
        & Instruction & 91.0 & 71.6 & 5.1 & 4.9 & & 82.9 & 52.1 & 15.1 & 4.9 \\
        & Steering (2-only) & 90.9 & \textbf{76.9 \gap{+5.3}} & \textbf{4.1 \gap{-1.0}} & 4.9 & & 83.1 & 59.5 & 21.2 & 4.9 \\
        & Steering (2+3) & 91.2 & 77.0 & 4.2 & 4.9 & & 85.9 & 59.7 & 15.5 & 4.9 \\
        & \cellcolor{green!5}Hybrid & \cellcolor{green!5}\textbf{92.2 \gap{+1.2}} & \cellcolor{green!5}76.3 & \cellcolor{green!5}4.4 & \cellcolor{green!5}4.9 &\cellcolor{green!5} & \cellcolor{green!5}\textbf{87.9 \gap{+5.0}} & \cellcolor{green!5}\textbf{62.9 \gap{+10.8}} & \cellcolor{green!5}\textbf{12.4 \gap{-2.7}} & \cellcolor{green!5}4.9 \\
        \midrule
        
        \multirow{4}{*}{Qwen3-14B}
        & Instruction & 92.1 & 72.2 & 5.2 & 4.9 & & 88.2 & 61.4 & 11.2 & 4.9 \\
        & Steering (2-only) & 92.9 & 75.2 & 4.6 & 4.9 & & 90.4 & 68.0 & 13.9 & 4.9 \\
        & Steering (2+3) & 93.0 & 73.8 & 5.3 & 4.9 & & 89.7 & 63.9 & 17.3 & 4.9 \\
        & \cellcolor{green!5}Hybrid & \cellcolor{green!5}\textbf{93.9 \gap{+1.8}} & \cellcolor{green!5}\textbf{78.3 \gap{+6.1}} & \cellcolor{green!5}\textbf{2.7 \gap{-2.5}} & \cellcolor{green!5}4.9 &\cellcolor{green!5} & \cellcolor{green!5}\textbf{91.7 \gap{+3.5}} & \cellcolor{green!5}\textbf{69.2 \gap{+7.8}} & \cellcolor{green!5}\textbf{6.2 \gap{-5.0}} & \cellcolor{green!5}4.9 \\
        
        \bottomrule
    \end{tabular}
    }
        \caption{Scaling analysis comparing steering tokens, text instructions, and their hybrid combination. Both steering and instruction methods benefit from scale, while training on 2-behavior combinations proves sufficient: explicit 3-behavior supervision (2+3) helps variance at 8B but degrades performance at 14B, suggesting larger models learn compositional patterns from simpler examples. Hybrid methods achieve the best accuracy-variance tradeoff at all scales. Numbers in brackets---improvement on top ``Instruction'' baseline. See App.~\ref{app:extra-experiments} for extended analysis.}
    \label{tab:steering-scalling-larger-models}
\end{table*}

\subsection{Steering tokens are superior compositional generalizers}
\label{sec:steering-vs-baselines}

Table~\ref{tab:steering-vs-baselines} compares our compositional steering against the baselines---instructions, merging LoRA adapters \cite{da2025steering} and output steering \cite{han2024word}---for 2- and 3-behavior compositions, with Qwen3-8B as the LLM. We report performance for \textit{seen} (during \texttt{<and>} training) and \textit{unseen} behavior compositions, with the latter indicating compositional generalization.

We find that training an explicit composition operator (\texttt{<and>} token) is crucial: simple concatenation of behavior tokens or using the native ``and'' token as a compositional operator collapses on unseen 3-behavior composition (33.2\% and 47.1\% vs. 59.6\%) and exhibits larger order variance. 
Our full compositional steering outperforms text instructions on \textit{unseen} compositions (+5.1\% for 2-behavior compositions, +5.5\% for 3-behavior compositions), while performing comparably on \textit{seen} compositions. This implies that our learned composition operator generalizes better than natural language composition.
Text instructions are more order-robust, exhibiting lower order variance, despite lower accuracy.
Importantly, Hybrid steering (tokens + instructions) achieves the best overall results (62.9\% accuracy, 15.2\% variance for the most difficult generalization case: \textit{unseen} 3-behavior compositions. This encouraging result suggests complementarity between our compositional token steering and natural language guidance. 
LoRA DARE fails to generalize compositionally, performing much worse on \textit{unseen} compositions, while CAA and LM-Steer completely collapse.\footnote{LM-Steer applies a linear transformation to output word embeddings and is designed for soft distributional preferences (e.g., sentiment); it is a poor fit for our hard constraints (exact word counts, strict language, character-level formatting).}
Response quality remains solid and is largely comparable across steering methods; except for CAA and LM-Steer, which exhibit broken model outputs. 

\begin{table}[t]
    \centering
    \resizebox{1.0\columnwidth}{!}{
    \begin{tabular}{lc ccc}
        \toprule
        \textbf{\texttt{<and>} init.} & \textbf{$\mathcal{L}_{\text{orth}}$} & \textbf{Seen $\uparrow$} & \textbf{Unseen $\uparrow$} & \textbf{Ord. Var. $\downarrow$} \\ 
        \midrule
        No \texttt{<and>} token & -- & 73.6 & 49.7 & 27.0 \\ \midrule
        Zero vector & \xmark & \textbf{94.5} & 66.9 & 11.2 \\
        \rowcolor{green!5}Zero vector & \cmark & 93.7 \gap{-0.8} & \textbf{69.1} \gap{+2.2} & \textbf{6.2} \gap{-5.0} \\
        \midrule
        ``and'' embedding & \xmark & \textbf{94.2} & 55.2 & 9.4 \\
        \rowcolor{green!5}``and'' embedding & \cmark & \textbf{94.2} \gap{+0.0} & \textbf{70.8} \gap{+15.6} & \textbf{7.1} \gap{-2.3} \\
        \midrule
        Avg. steering tokens & \xmark & 94.3 & 58.4 & 8.6 \\
        \rowcolor{green!5}Avg. steering tokens & \cmark & \textbf{93.5} \gap{-0.8} & \textbf{64.4} \gap{+6.0} & \textbf{6.2} \gap{-2.4} \\
        \bottomrule
    \end{tabular}
    }
    \caption{Ablation study of \texttt{<and>} token initialization and orthogonality regularization (Qwen3-4B; 2-behavior composition). The initializations effect matters only on unseen combinations, isolating compositional generalization as the key differentiator. Orthogonality loss proves critical, while without the \texttt{<and>} token, performance collapses. Zero initialization with orthogonality provides the best accuracy-variance tradeoff.}
    \label{tab:and-token-ablation}
\end{table}

\begin{table}[t]
    \centering
    \resizebox{0.8\columnwidth}{!}{
    \begin{tabular}{l cc}
        \toprule
        \textbf{$\lambda$} & \textbf{Unseen(2|3) $\uparrow$} & \textbf{Ord. Var. (2|3) $\downarrow$} \\ 
        \midrule
        0.0 & 67.6 | 59.8 & 5.3 | 21.7 \\
        0.25 & 65.0 | 53.4 & 5.1 | \textbf{21.1} \\
        \rowcolor{green!5}0.5 (default) & \textbf{69.1} | \textbf{60.7} & 6.2 | 22.7 \\
        1.0 & 63.9 | 51.3 & \textbf{5.0} | 22.3 \\
        \bottomrule
    \end{tabular}
    }
    \caption{Ablation study of the orthogonality loss weight ($\lambda$) using Qwen3-4B on 2|3 unseen behavior composition. $\lambda = 0.5$ performs best and $\lambda=1.0$ worst. Order variance shows no clear trend. Overall, the method does not seem to be overly sensitive to the choice of $\lambda$.}
    \label{tab:lambda-ablation}
\end{table}


\subsection{Robustness across model families}\label{sec:steering-different-models}
Table~\ref{tab:steering-different-models} summarizes the compositional steering performance for steering tokens, instructions, and their Hybrid across seven models (Qwen3-4B/8B, Llama3-3B/8B, SmolLM3-3B, OLMo-7B, Gemma3-4B) and five model families. 
The results render our findings from \S\ref{sec:steering-vs-baselines} to be robust across different LLM architectures: (1) our steering tokens largely outperform instruction-steering (and when not, they offer on par performance); (2) the Hybrid combination of tokens and instructions offers further gains. There is, however, some family-based variance: Qwen models are more steerable, and favor compositional steering tokens over instruction steering (76.9\% for unseen 2-behavior on Qwen3-8B); Llama models, in contrast, are much less steerable, regardless of the steering approach (Llama3-8B: 39.5\% token steering, 40.2\% instructions). Importantly, the Hybrid combination of token-based compositional steering and natural language instructions rescues weak models: we observe a +12.7\% gain for Llama3-8B for 3-behavior compositions. This again points to strong complementarity between our compositional token steering and natural language instructions. 

\subsection{Robustness across model sizes}\label{sec:steering-scalling-larger-models}
We next investigate whether compositional accuracy and robustness of our compositional steering tokens scale with model size, as well as whether training on both 2- and 3-behavior compositions improves generalization (compared to our default training on 2-behavior compositions only). Table~\ref{tab:steering-scalling-larger-models} summarizes the results. 
We observe that both compositional token steering and instruction-based steering benefit from scale. 
Performance on 3-behavior compositions for compositional steering tokens (training on 2-behavior compositions only) jumps from 59.5\% at 8B to 68.0\% at 14B (+8.5\%), while instructions improve from 52.1\% to 61.4\% (+9.3\%). The Hybrid approach also scales effectively, achieving 69.2\% at 14B and becomes much more robust to behavior ordering, exhibiting only 6.2\% order variance (compared to 18.6\% at 4B). Somewhat surprisingly, explicitly training our composition token also for 3-behavior compositions degrades performance at 14B: (2-behavior +3-behavior training yields only 63.9\% accuracy, compared to 68.0\% we get when training only on 2-behavior compositions; and it also exhibits higher order variance (17.3\% vs. 13.9\%). This again suggests that we successfully learned a general composition operator (i.e., already from the 2-behavior compositions).
At 8B, 2- + 3-behavior training provides marginal variance reduction (compared to 4B; from 21.2\% to 15.5\%) with comparable accuracy, indicating that training data efficiency is likely scale-dependent. 
Note that, in Appendix~\ref{app:four-behavior} we perform inference using Qwen3-14B on 4-behavior compositions where the 4-th behavior is \textit{json wrap} of the LLM outputs. We find that at 4-behavior compositions, pure steering becomes order-unstable and hybrid is the regime that scales.

\subsection{Ablation studies: Effect of \texttt{<and>} token initialization and $\lambda$ weight}\label{sec:and-token-ablation}

In Table~\ref{tab:and-token-ablation}, we ablate initialization strategies
and orthogonality regularization, to isolate what mechanisms enable compositional learning in the \texttt{<and>} token. We report results using Qwen3-4B.

We first observe that \texttt{<and>} is essential: without it, \textit{seen} performance degrades to 73.6\% (vs.\,93-95\% with \texttt{<and>}), and \textit{unseen} to 49.7\% (vs.\,64-71\%), and order variance increases to 27.0\% (vs. 5-11\%).
All \texttt{<and>} initialization strategies achieve comparable \textit{seen} performance (93-95\%), suggesting that behavior learning does not depend on initialization. However, \textit{unseen} performance varies widely (55-71\%), isolating compositional generalization as the key differentiator. Orthogonality regularization is critical here, especially for semantic behavior initialization: ``and'' embedding without orthogonality achieves only 55.2\% \textit{unseen}, but with the orthogonality constraint it improves by 15.6\%. For zero initialization, orthogonality primarily reduces variance from 11.2\% to 5.3\% while only modestly improving accuracy.
Token-average initialization benefits least from orthogonality, suggesting that averaging of behavior embeddings provides a poor starting point for composition. 
These results stress the importance of (i) an explicit learned operator (the \texttt{<and>} token), and (ii) enforcing the orthogonality of its embedding to behavior token representations. 

We further conduct a $\lambda$ ablation on Qwen3-4B (zero initialization, 2-behavior training). We report results in Table~\ref{tab:lambda-ablation}. We observe that $\lambda=0.5$ performs best (69.1\% 2-tok, 60.7\% 3-tok) and $\lambda=1.0$ worst (63.9\%, 51.3\%). Order variance shows no clear trend across values of $\lambda$. We conclude that the method does not seem to be overly sensitive to the choice of this hyperparameter. Finally, see Appendix~\ref{app:interpretability} for an interpretability analysis of the \texttt{<and>} token.

\subsection{Per-behavior breakdown}\label{sec:steering-behavior-breakdown}

\begin{figure*}[t]
    \centering
    \begin{subfigure}[t]{0.24\textwidth}
        \centering
        \includegraphics[width=\linewidth]{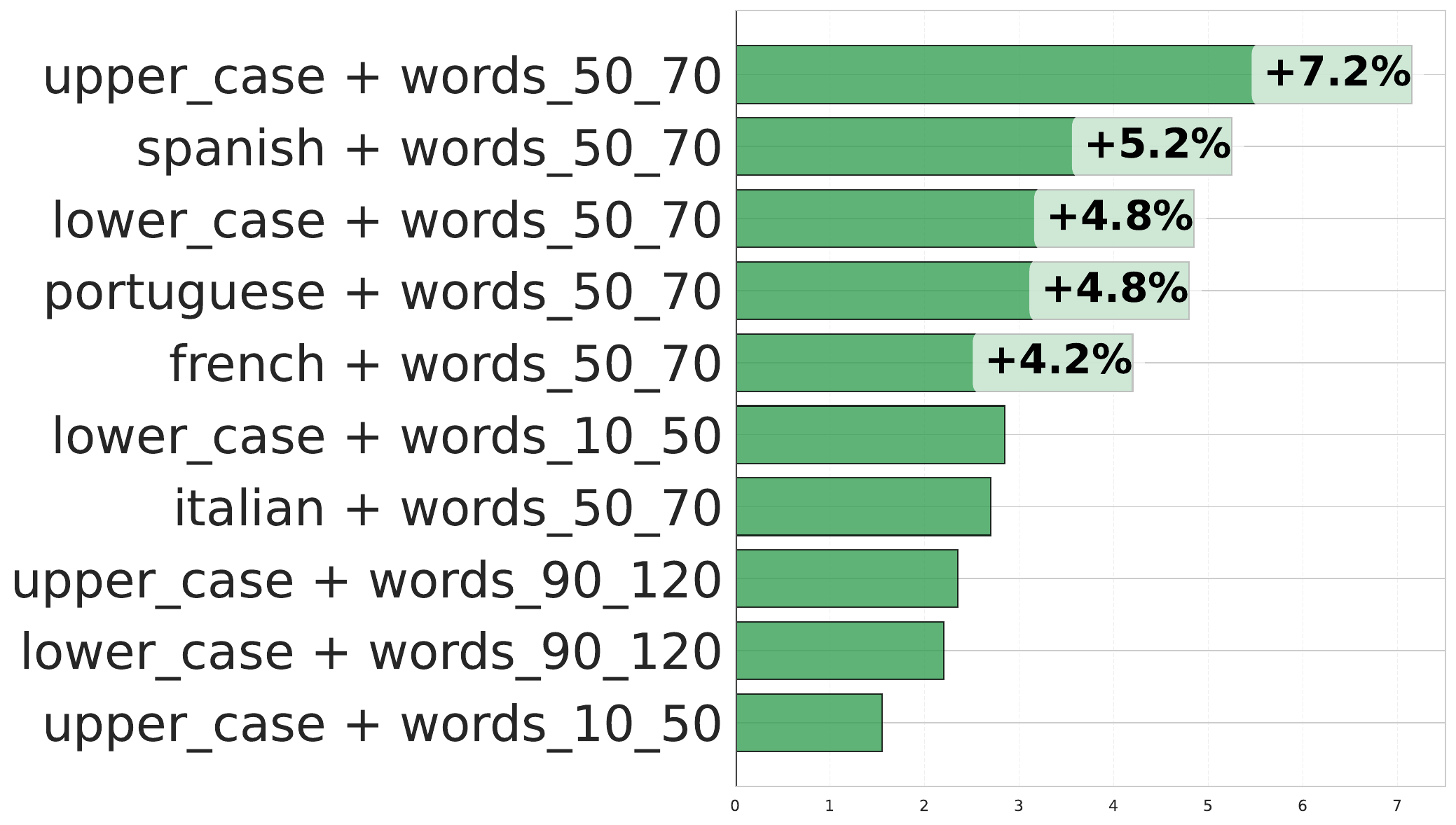}
        \label{fig:seen-steer-qwen}
    \end{subfigure}\hfill
    \begin{subfigure}[t]{0.24\textwidth}
        \centering
        \includegraphics[width=\linewidth]{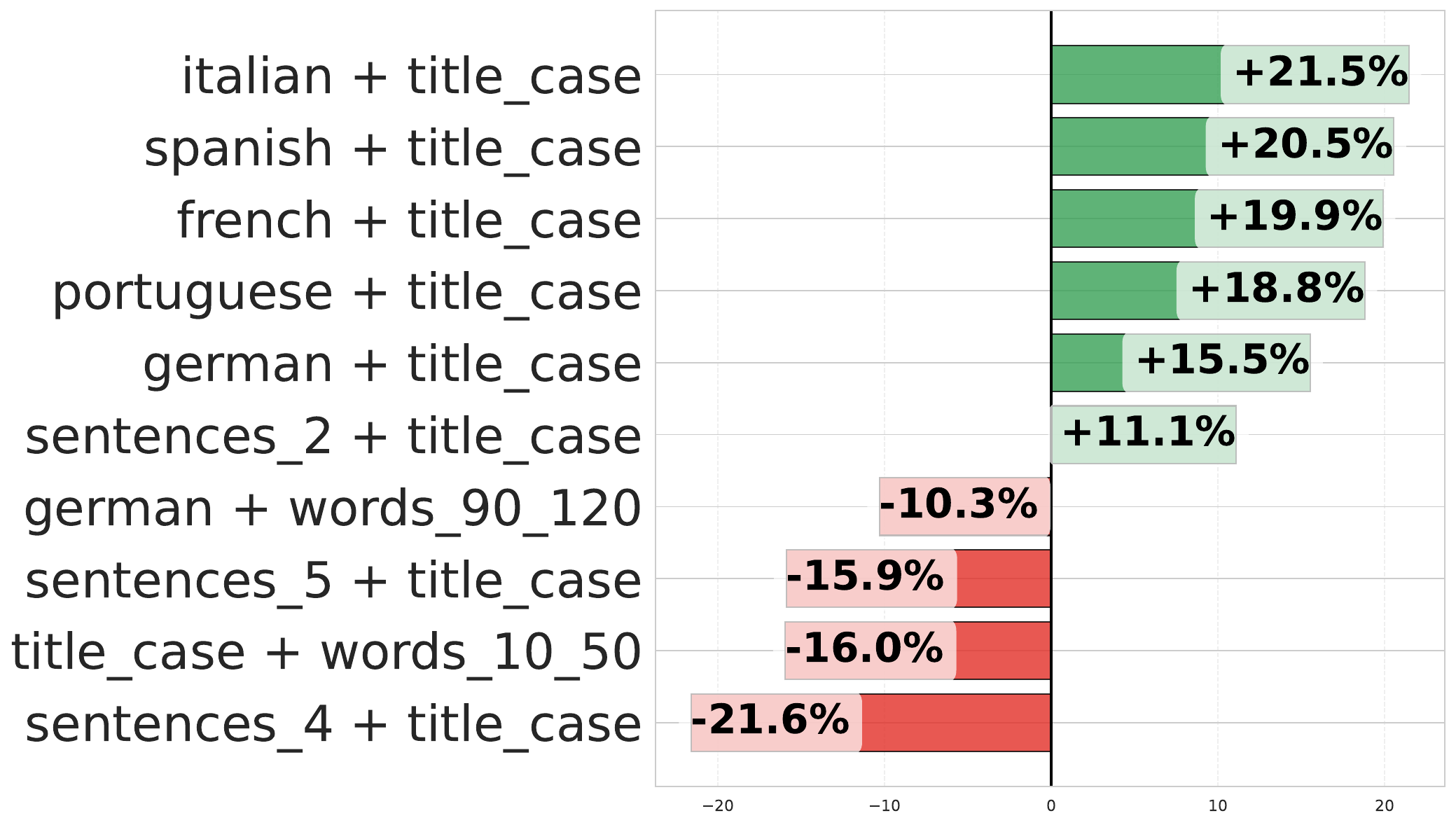}
        \label{fig:unseen-steer-qwen}
    \end{subfigure}\hfill
    \begin{subfigure}[t]{0.24\textwidth}
        \centering
        \includegraphics[width=\linewidth]{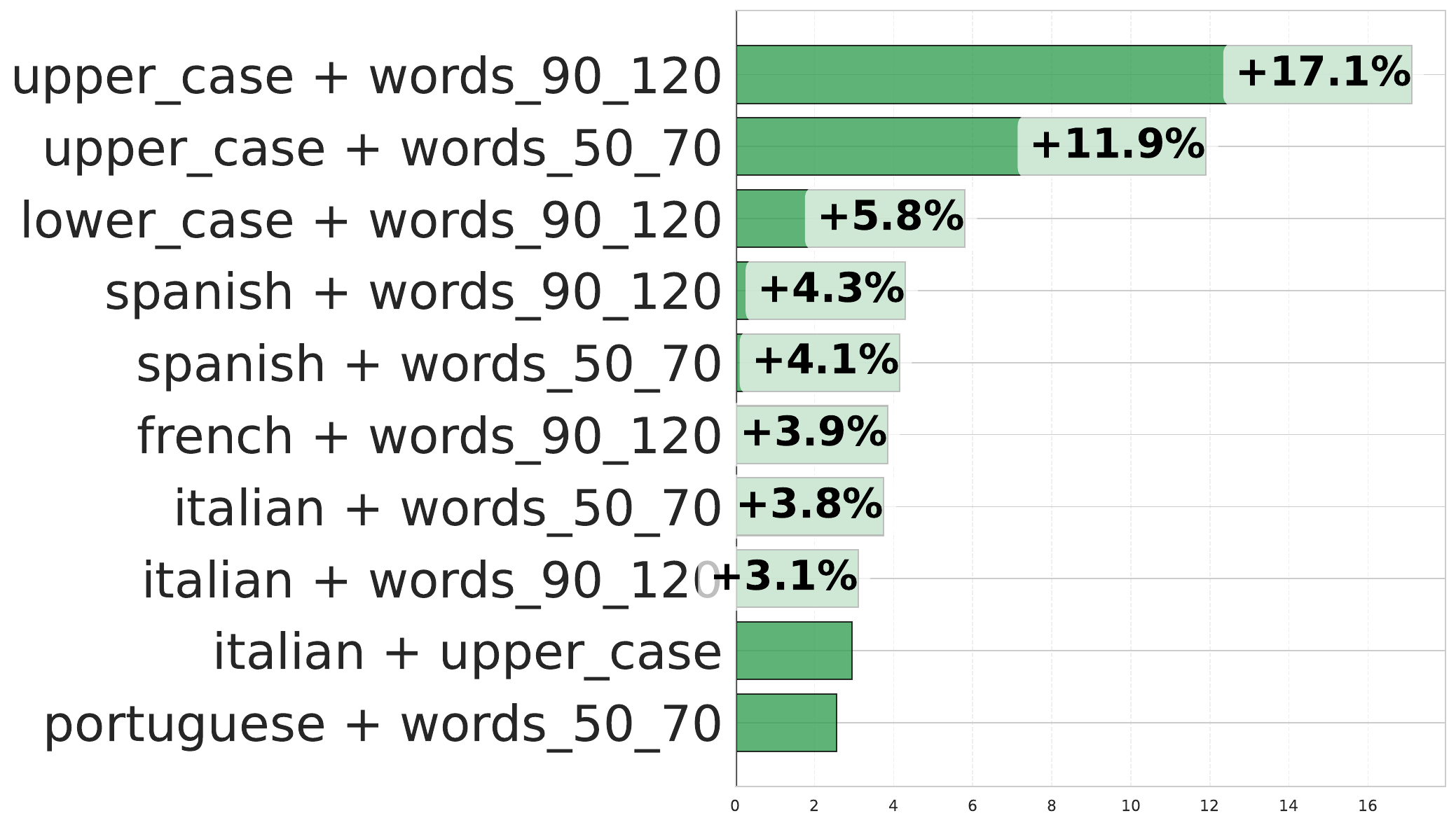}
        \label{fig:seen-hybrid-qwen}
    \end{subfigure}\hfill
    \begin{subfigure}[t]{0.24\textwidth}
        \centering
        \includegraphics[width=\linewidth]{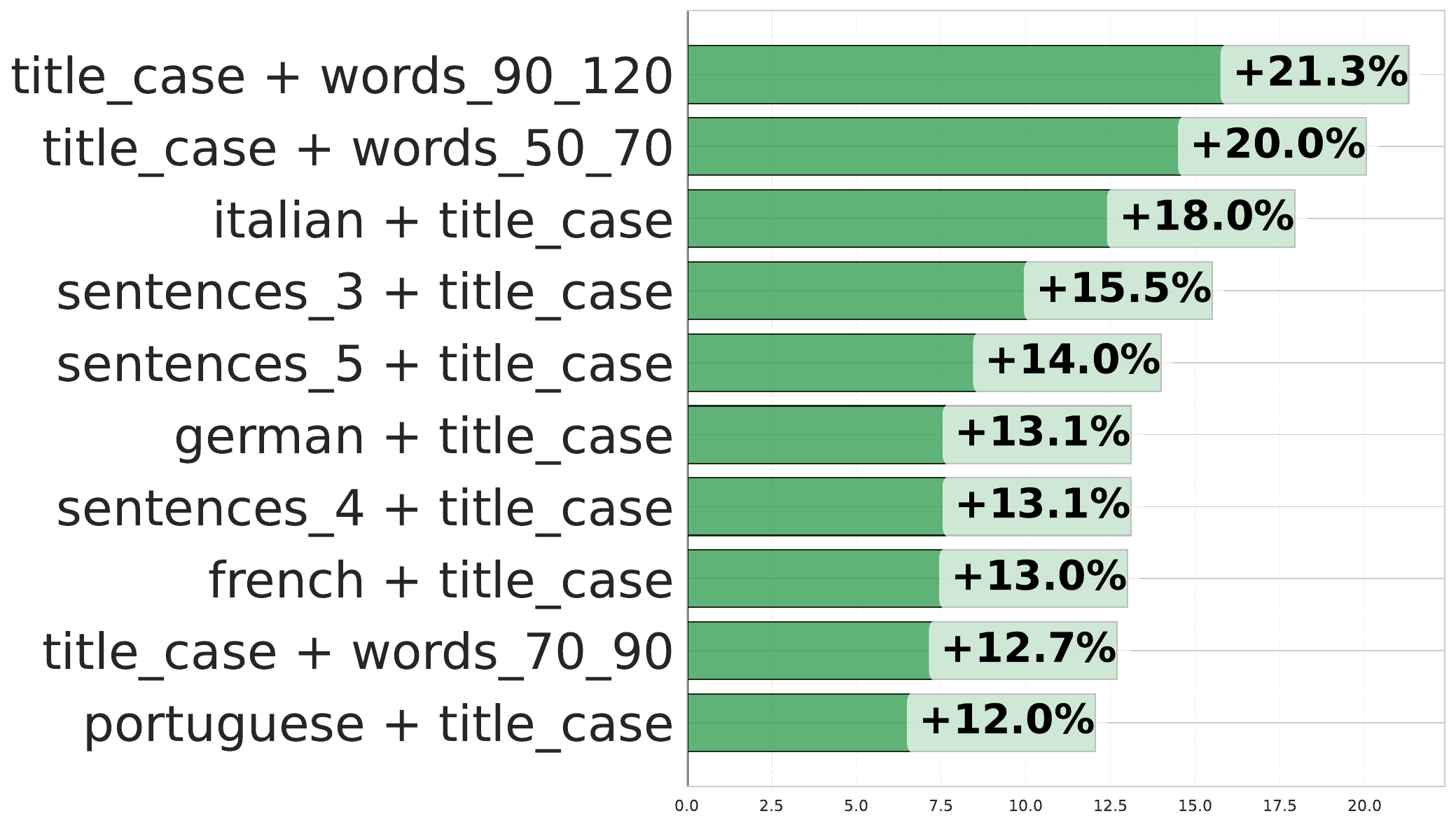}
        \label{fig:unseen-hybrid-qwen}
    \end{subfigure}

    \vspace{-1.0em}

    \begin{subfigure}[t]{0.24\textwidth}
        \centering
        \includegraphics[width=\linewidth]{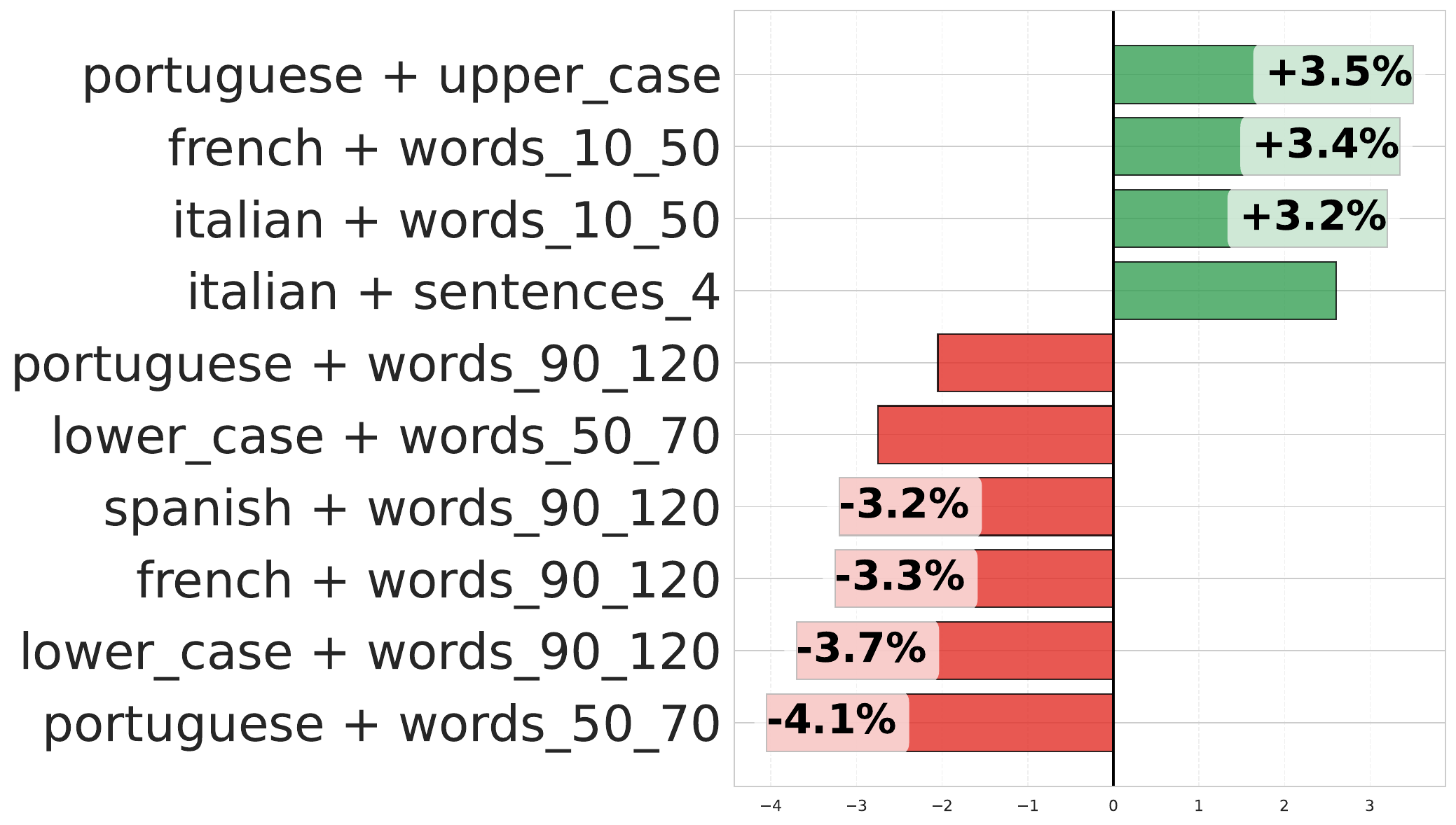}
        \caption{Seen – Steering vs Text}
        \label{fig:seen-steer-llama}
    \end{subfigure}\hfill
    \begin{subfigure}[t]{0.24\textwidth}
        \centering
        \includegraphics[width=\linewidth]{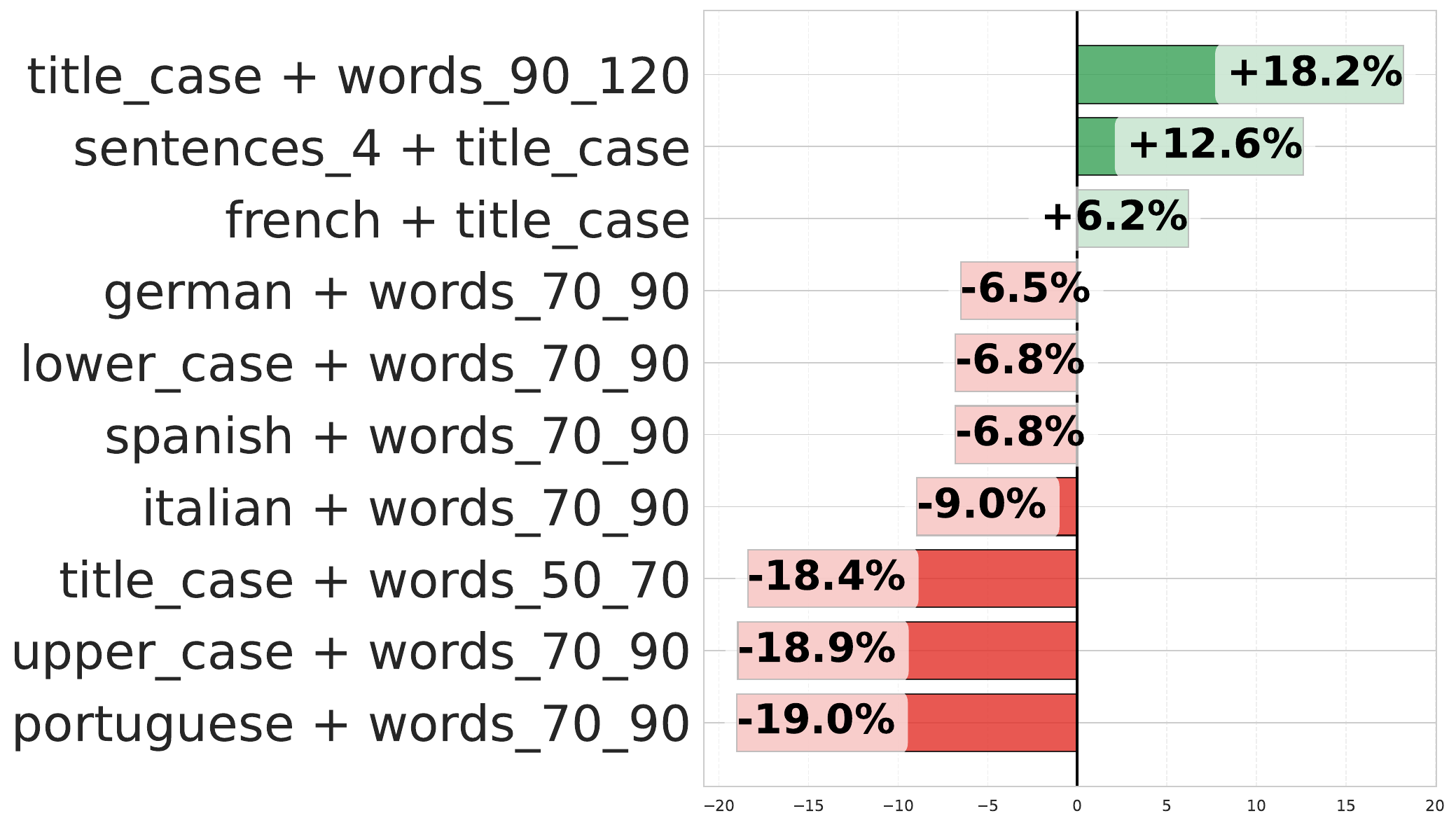}
        \caption{Unseen – Steering vs Text}
        \label{fig:unseen-steer-llama}
    \end{subfigure}\hfill
    \begin{subfigure}[t]{0.24\textwidth}
        \centering
        \includegraphics[width=\linewidth]{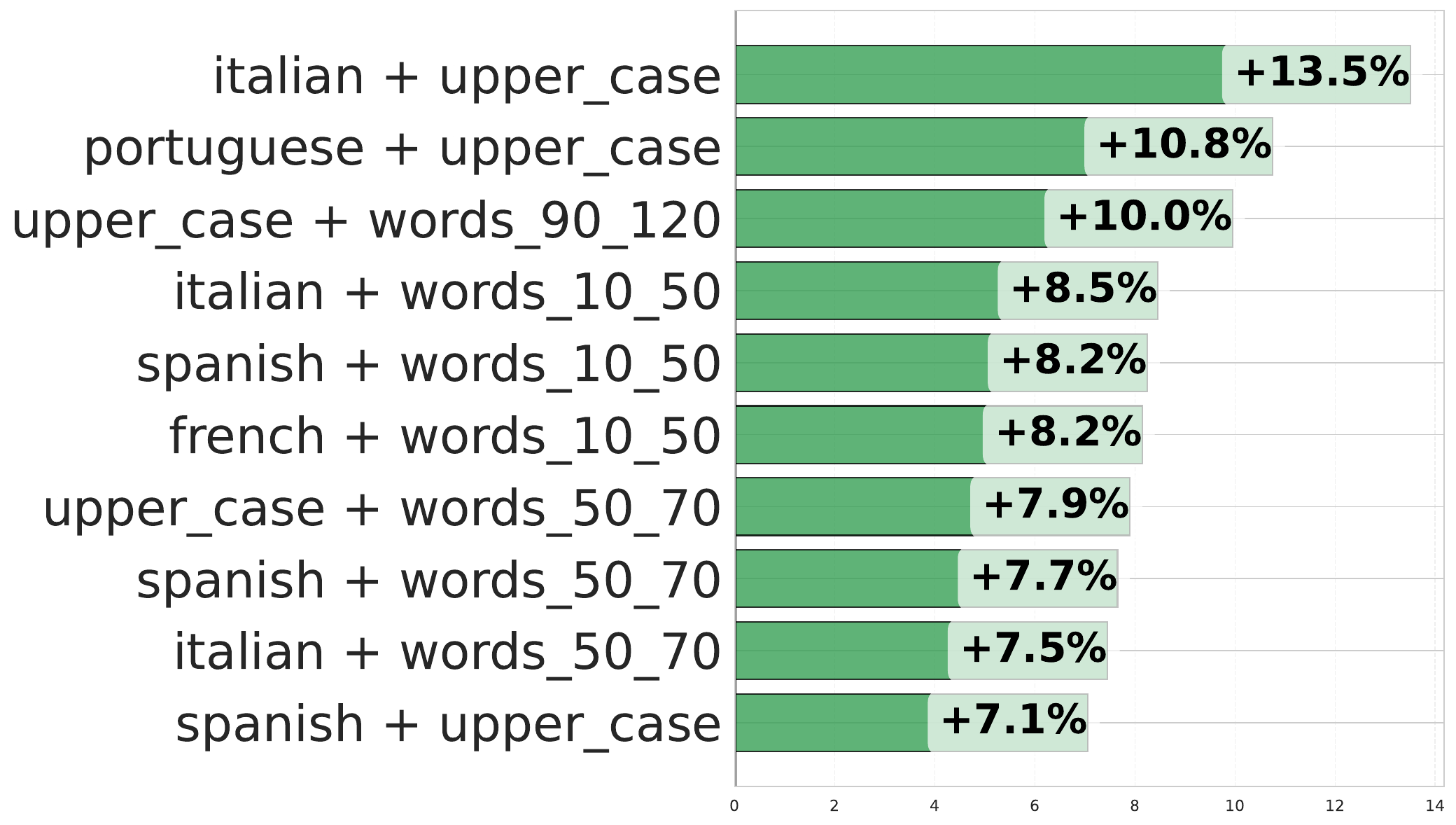}
        \caption{Seen – Hybrid vs Text}
        \label{fig:seen-hybrid-llama}
    \end{subfigure}\hfill
    \begin{subfigure}[t]{0.24\textwidth}
        \centering
        \includegraphics[width=\linewidth]{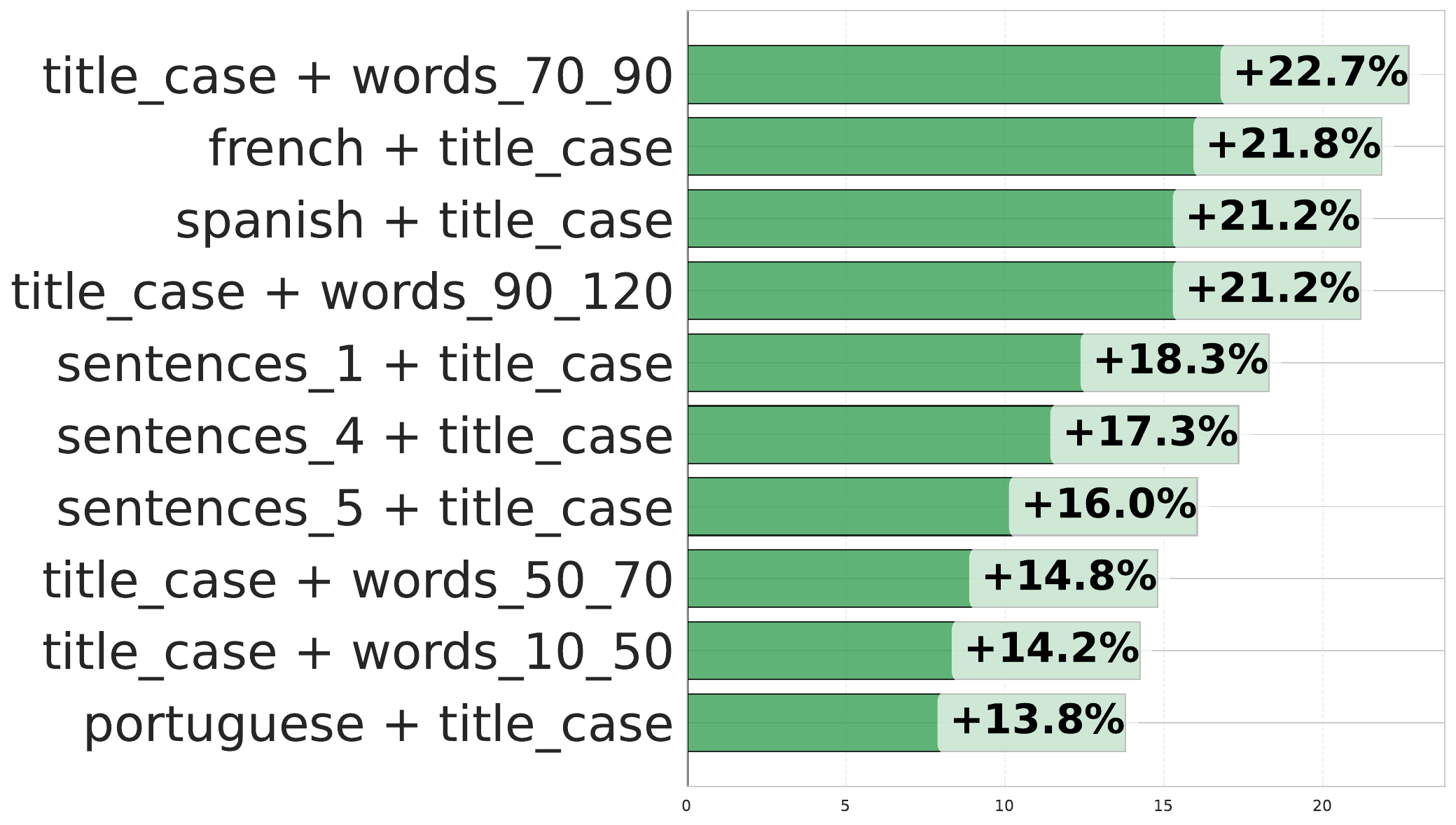}
        \caption{Unseen – Hybrid vs Text}
        \label{fig:unseen-hybrid-llama}
    \end{subfigure}

    \caption{Average relative performance gains (\%) for Qwen3-14B (top row) and Llama3-8B (bottom row). Top 20 behavior combinations (2- and 3-behavior) yielding the largest differences between methods. Green bars indicate improvements over the text baseline; red bars indicate degradations.}
    \label{fig:steering-behavior-breakdown}
\end{figure*}

Figure~\ref{fig:steering-behavior-breakdown} presents a granular per-composition analysis for Qwen3-14B and Llama3-8B, revealing which behaviors benefit most from compositional token steering vs. text instructions. We find that unseen behavior combinations, and in particular \texttt{title\_case} paired with language or length constraints, drive the gains of token steering. This indicates that unseen \textit{cross-category} compositions benefit the most from learned \texttt{<and>} operators. However, LLM-dependent failure modes emerge: while Qwen3-14B shows better performance with token steering for most unseen combinations, Llama3-8B exhibits stark category-specific divergence, with token steering excelling on compositions with formatting behaviors but failing on compositions with the \textit{unseen} length behavior (\texttt{words\_70\_90}), where the instruction steering seems superior. Critically, the hybrid combination seems to eliminate these failure modes, reaffirming that complementarity between steering with learned embeddings and text instructions is key for robust compositional control, especially for weaker LLMs.


\section{Conclusion}\label{sec:conclusion}

In this work, we presented \textit{steering tokens}, an effective approach for compositional control of LLMs, that does not require modification of the models' internals. By means of self-distillation, in the first step we train input embeddings for individual (automatically verifiable) model behaviors, which we call steering tokens. In the second step, we freeze the behavior-specific steering tokens and train only the explicit composition token \texttt{<and>}. Our extensive experimentation renders steering tokens (1) superior to instruction-based steering as well as to other steering methods, while maintaining comparable response quality. Further, we show that the steering tokens are  (2) complementary to instructions, obtaining further gains from combination with natural language-based guidance. Crucially, (3) we demonstrate that steering tokens successfully generalize to behaviors and compositions \textit{unseen} during the training of our composition operator (i.e., the \textit{<and>} token). 



Finally, we show that our compositional steering tokens generalize well across different model families and sizes, provide even larger steering gains for larger models. Importantly, training exclusively on 2-behavior compositions proves sufficient for larger models. 
%
%
This work introduces steering tokens as a solid
mechanism for compositional control in LLMs, offering an efficient alternative to instruction-based steering while complementing it.

\section*{Limitations}

While steering tokens demonstrate strong compositional capabilities, several limitations warrant attention in future work:

\textbf{Constraint verifiability.} Our evaluation focuses exclusively on verifiable behavioral properties (response length, formatting conventions, etc.) where ground-truth satisfaction is automatically assessed. Steering tokens represent a general control mechanism that applies to broader semantic constraints (tone, style, etc.), but evaluating such properties requires human annotation or LLM-based judges. Future work should extend compositional evaluation to subjective and nuanced behavioral dimensions.

\textbf{Compositional complexity.} We evaluate compositions of up to three properties, demonstrating zero-shot generalization from 2-property training to 3-property inference. Real-world applications may require simultaneous control over many more constraints (e.g., language, length, tone, domain, etc.). Therefore, scaling steering tokens to higher-order compositions remains an open question, particularly regarding whether compositional accuracy degrades gracefully or rapidly as the property count increases.

\textbf{Model scale.} Our experiments span 3B to 14B parameter models, observing compositional improvements with scale. However, frontier models now exceed those sizes, and it remains unclear whether steering tokens continue to benefit from scale or encounter diminishing returns. Extension to larger models would clarify whether compositional reasoning improves further and whether hybrid methods remain necessary at scale.
\section*{Acknowledgements}
We thank Andreas Ripke for his support with hosting the LLMs used in our experiments. The work of Goran Glavaš was supported by the Alcatel-Lucent Stiftung and Deutsches Stiftungszentrum through the grant ``Equitably Fair and Trustworthy Language Technology'' (EQUIFAIR, Grant Nr. T0067/43110/23).
\clearpage

\bibliography{custom}
\clearpage
\appendix
\section*{Supplementary Matrial}
We organize the supplementary material as follows:

\begin{itemize}
    \item \S\ref{app:soft-prompt-tuning}: Discussion how our method is related to soft prompt tuning \cite{li2021prefix}.
    \item \S\ref{app:implementation-details}: All implementation details in order to ensure reproducibility of our work.
    \item \S\ref{app:four-behavior}: Experiments featuring 4-behavior composition.
    \item \S\ref{app:interpretability}: Interpretability analysis of the steering tokens.
    \item \S\ref{app:extra-experiments}: Additional experiments that we did not include in the main paper.
    \item \S\ref{app:llm-as-judge}: The LLM-as-judge prompts we use to measure the response quality of the outputs.
    \item \S\ref{app:metrics}: Additional details about the evaluation metrics.
    \item \S\ref{app:baselines}: Implementation details for the baselines.
    \item \S\ref{app:instruction-variants}: The instructions we use for each steering behavior.
    \item \S\ref{app:generations}: Model outputs for different steering combinations.
\end{itemize}

\section{``Behavior Steering'' vs.\ Soft/Postfix Tuning}\label{app:soft-prompt-tuning}

Our method is related to soft prompt tuning~\citep{lester2021power,li2021prefix}, and since our tokens are appended after the prompt, ``postfix tuning'' is an accurate description of the mechanism. There is, however, one crucial difference w.r.t.\ standard soft prompting: our interventions are strictly limited to the input (i.e., the addition of steering tokens), whereas soft prompting (prefix or postfix tuning), when used as a parameter-efficient fine-tuning approach (PEFT), typically introduces trainable parameters in each transformer layer -- and usually significantly more than a single trainable embedding vector per fine-tuning task. It is precisely the fact that we manipulate only the input space that allows for compositional generalization, i.e., to steer the model ad-hoc at inference for behavior combinations not seen in training; this is not possible for PEFT approaches that modify model internals (as shown in Table~\ref{tab:steering-vs-baselines} for the LoRA-based baseline).

However, the key distinction is not the training mechanism, but rather the compositional generalization objective. Soft prompt tuning optimizes a fixed embedding for a single task. In contrast, we introduce a dedicated composition operator (the \texttt{<and>} token) trained on behavior pairs, which then generalizes zero-shot to: (i) unseen behavior combinations, (ii) combinations involving behaviors never seen during \texttt{<and>} training, and (iii) compositions with some behaviors never seen at training (3-behavior and 4-behavior compositions). This generalization across behavior combinations is neither demonstrated by nor the goal of prior soft prompt tuning methods.

We adopt the term ``steering'' in the sense established by the behavior steering literature (controlling specific output properties), which is orthogonal to the (input) placement of the steering tokens. Our key novelty is not the training mechanism (input-space optimization with a frozen LLM) but the compositional generalization enabled by the learned \texttt{<and>} operator---a property not demonstrated by prior work. Finally, we note that our method requires LLM gradients during training (similar to other steering methods, e.g., LM-Steer~\citep{han2024word}), unlike activation-steering methods that compute steering vectors as aggregates of activations.

\section{Implementation Details}\label{app:implementation-details}
We train all steering tokens using AdamW with learning rate $\alpha = 1 \times 10^{-4}$, weight decay $\lambda = 1 \times 10^{-3}$, batch size as large as we can fit on a GPU, and 2 epochs over 50,000 training examples. Learning rate follows linear warmup (10\% of steps) then linear decay to 0. Gradients are clipped to a maximum norm of 1.0. We set the temperature to 10.0. We use bfloat16 mixed precision training. Training data is preprocessed to remove the top 0.1\% longest examples.

\section{4-way Behavior Composition}\label{app:four-behavior}
We train a new 4th behavior steering token, in order to test for 4-way compositions of behaviours via the \texttt{<and>} token, which remains trained on 2-behavior compositon only. We denote the behavior as \textit{json}, where we steer the model towards having the answer in JSON format, where the key ``answer'' is mandatory. We evaluate for this by checking if (i) the JSON can be parsed via \texttt{json.loads} in Python; (ii) the JSON object contains the ``answer'' as key; and (iii) has only one key. We train this behavior for Qwen3-14B, and pair it with the \textit{languages}, \textit{formatting}, \textit{length}, and \textit{structure}, and report the results in Table~\ref{tab:four-behavior}.

\begin{table}[t]
    \centering
    \resizebox{1.0\columnwidth}{!}{
    \begin{tabular}{l cc}
        \toprule
        \textbf{Method} & \textbf{Acc. (3|4 Behavior) $\uparrow$} & \textbf{Ord. Var. (3|4) $\downarrow$} \\ 
        \midrule
        Instr. steering \cite{stolfo2024improving} & 70.6 | 64.4 & 13.7 | \textbf{19.5} \\
        Composition (\texttt{<and>}) & 76.3 | 53.5 & 14.1 | 60.0 \\
        \rowcolor{green!5}Hybrid & \textbf{78.0} | \textbf{67.5} & \textbf{7.0} | 25.5 \\
        \bottomrule
    \end{tabular}
    }
    \caption{4-behavior composition using \texttt{<and>} steering token trained with only 2-behavior data.}
    \label{tab:four-behavior}
\end{table}

We observe that the \texttt{<and>} token does generalize to 4-behavior compositions -- non-trivially, since it has never seen $2 \rightarrow 4$ in training. Pure steering's order variance increases substantially at 4-behaviors (60.0\%), while the hybrid method keeps it more contained (25.5\%), though instruction steering remains the most order-robust (19.5\%). We consider the fact that the \texttt{<and>} token generalizes across behaviors ($2 \rightarrow 3 \rightarrow 4$) with a monotonic but contained drop is strong evidence of a genuine compositional operator rather than a single $2 \rightarrow 3$ jump.

\section{Additional Results}\label{app:extra-experiments}
We report supplementary results to the ones in the main paper in Table~\ref{tab:steering-different-models-best}, Table~\ref{tab:steering-scalling-larger-models-best}, and Table~\ref{tab:single-token-performance}.

\begin{table*}[t]
    \centering
    \resizebox{1.0\textwidth}{!}{
    \begin{tabular}{ll cccccc}
        \toprule
         Model & Method & \makecell{2-Tok Seen \\ Mean / Best ($\uparrow$)} & \makecell{2-Tok Unseen \\ Mean / Best ($\uparrow$)} & \makecell{2-Tok Var \\ Both ($\downarrow$)} & \makecell{3-Tok Seen \\ Mean / Best ($\uparrow$)} & \makecell{3-Tok Unseen \\ Mean / Best ($\uparrow$)} & \makecell{3-Tok Var \\ Both ($\downarrow$)} \\ \midrule
         \multirow{3}{*}{Qwen3-4B}
         & Instruction & 93.3 / 94.7 & 68.9 / 73.2 & \textbf{5.3} & 88.2 / 91.7 & 55.6 / 64.8 & \textbf{15.1} \\
         & Steering token & \textbf{93.7} / 94.7 & 69.1 / \textbf{75.0} & 6.2 & 89.3 / 93.7 & \textbf{60.7 / 75.5} & 22.7 \\
         & Steering token + Instruction & \textbf{93.7} / \textbf{94.9} & \textbf{69.2} / 73.9 & \textbf{5.3} & \textbf{90.7 / 93.5} & 58.0 / 71.2 & 18.6 \\ \midrule
         \multirow{3}{*}{Qwen3-8B}
         & Instruction & 90.7 / 92.2 & 71.8 / 75.7 & 5.1 & 83.7 / 88.7 & 54.0 / 63.9 & 15.1 \\
         & Steering token & 90.9 / 92.5 & \textbf{76.9 / 79.6} & 4.1 & 83.1 / 89.8 & 59.5 / 70.7 & 21.2 \\
         & Hybrid & \textbf{92.2 / 93.7} & 76.3 / 79.6 & \textbf{4.4} & \textbf{87.9 / 91.8} & \textbf{62.9 / 70.3} & \textbf{12.4} \\ \midrule
         \multirow{3}{*}{Llama3-3B}
         & Instruction & 86.6 / 87.7 & 66.7 / 68.9 & 3.1 & 68.5 / 72.0 & 33.8 / 38.7 & \textbf{8.7} \\
         & Steering token & 87.7 / 89.1 & 69.3 / 72.0 & 3.9 & 69.8 / 74.5 & 33.9 / 45.8 & 18.7 \\
         & Hybrid & \textbf{91.7 / 92.4} & \textbf{74.9 / 77.3} & \textbf{2.8} & \textbf{76.9 / 79.9} & \textbf{43.4 / 50.4} & 10.6 \\ \midrule
         \multirow{3}{*}{Llama3-8B}
         & Instruction & 88.1 / 89.3 & 67.8 / 70.3 & 3.6 & 79.4 / 83.8 & 40.2 / 48.2 & 13.6 \\
         & Steering token & 88.1 / 89.3 & 67.0 / 72.4 & 5.9 & 80.7 / 84.2 & 39.5 / 54.1 & 19.2 \\
         & Hybrid & \textbf{92.3 / 93.2} & \textbf{76.3 / 78.8} & \textbf{3.2} & \textbf{86.9 / 89.0} & \textbf{52.9 / 61.3} & \textbf{11.0} \\ \midrule
         \multirow{3}{*}{Smol3-3B}
         & Instruction & 74.2 / 76.9 & 53.2 / 55.6 & \textbf{5.1} & 62.3 / 68.1 & 32.5 / 40.4 & \textbf{14.5} \\
         & Steering token & 68.0 / 73.0 & 53.2 / \textbf{60.4} & 11.8 & 50.3 / 65.9 & 35.5 / 53.0 & 35.9 \\
         & Hybrid & \textbf{75.7 / 79.1} & \textbf{53.5} / 57.5 & 7.3 & \textbf{66.2 / 73.9} & \textbf{37.2 / 46.5} & 17.1 \\ \midrule
         \multirow{3}{*}{Olmo-7B}
         & Instruction & 78.0 / 79.0 & 56.8 / 58.7 & \textbf{2.9} & 55.5 / 58.6 & 30.9 / 34.8 & 7.3 \\
         & Steering token & 77.5 / 78.5 & 56.9 / 59.7 & 3.6 & 53.9 / 56.8 & 28.4 / 36.6 & 12.3 \\
         & Hybrid & \textbf{78.3 / 80.2} & \textbf{60.9 / 62.7} & 3.7 & \textbf{57.8 / 60.8} & \textbf{37.5 / 40.6} & \textbf{6.4} \\
        \bottomrule
    \end{tabular}
    }
    \caption{Extension using \textit{best accuracy} of Table~\ref{tab:steering-different-models}.}
    \label{tab:steering-different-models-best}
\end{table*}

\begin{table*}[t]
    \centering
    \resizebox{1.0\textwidth}{!}{
    \begin{tabular}{ll ccccccc}
        \toprule
         Model & Method & \makecell{2-Tok Seen \\ Mean / Best ($\uparrow$)} & \makecell{2-Tok Unseen \\ Mean / Best ($\uparrow$)} & \makecell{2-Tok Var \\ Both ($\downarrow$)} & \makecell{3-Tok Seen \\ Mean / Best ($\uparrow$)} & \makecell{3-Tok Unseen \\ Mean / Best ($\uparrow$)} & \makecell{3-Tok Var \\ Both ($\downarrow$)} & \makecell{Response \\ Quality ($\uparrow$)} \\ \midrule
         \multirow{3}{*}{Qwen3-4B}
         & Instruction & 93.3 / 94.7 & 68.9 / 73.2 & \textbf{5.3} & 88.2 / 91.7 & 55.6 / 64.8 & \textbf{15.1} & 4.8 \\
         & Steering token (2-only) & 93.7 / 94.7 & 69.1 / \textbf{75.0} & 6.2 & 89.3 / 93.7 & \textbf{60.7 / 75.5} & 22.7 & 4.8 \\
         & Steering token + Instruction & \textbf{93.7 / 94.9} & \textbf{69.2} / 73.9 & \textbf{5.3} & \textbf{90.7 / 93.5} & 58.0 / 71.2 & 18.6 & 4.8 \\ \midrule
         \multirow{3}{*}{Qwen3-8B}
         & Instruction & 91.0 / 92.5 & 71.6 / 75.2 & 5.1 & 82.9 / 88.2 & 52.1 / 61.5 & 15.1 & 4.9 \\
         & Steering token (2-only) & 90.9 / 92.5 & 76.9 / 79.6 & \textbf{4.1} & 83.1 / 89.8 & 59.5 / \textbf{70.7} & 21.2 & 4.9 \\
         & Steering token (2+3) & 91.2 / 92.5 & \textbf{77.0 / 80.1} & 4.2 & 85.9 / 90.2 & 59.7 / 68.7 & 15.5 & 4.9 \\
         & Steering token + Instruction & \textbf{92.2 / 93.7} & 76.3 / 79.6 & 4.4 & \textbf{87.9 / 91.8} & \textbf{62.9} / 70.3 & \textbf{12.4} & 4.9 \\ \midrule
         \multirow{4}{*}{Qwen3-14B}
         & Steering token (2-only) & 92.9 / 94.5 & 75.2 / 78.4 & 4.6 & 90.4 / \textbf{94.3} & 68.0 / 76.0 & 13.9 & 4.9 \\
         & Steering token (2+3) & 93.0 / 94.6 & 73.8 / 77.9 & 5.3 & 89.7 / 93.1 & 63.9 / 73.7 & 17.3 & 4.9 \\
         & Instruction & 92.1 / 93.7 & 72.2 / 76.2 & 5.2 & 88.2 / 91.6 & 61.4 / 67.6 & 11.2 & 4.9 \\
         & Steering token + Instruction & \textbf{93.9 / 94.8} & \textbf{78.3 / 80.2} & \textbf{2.7} & \textbf{91.7} / 93.6 & \textbf{69.2} / \textbf{72.9} & \textbf{6.2} & 4.9 \\
        \bottomrule
    \end{tabular}
    }
    \caption{Extension using \textit{best accuracy} of Table~\ref{tab:steering-scalling-larger-models}.}
    \label{tab:steering-scalling-larger-models-best}
\end{table*}

\begin{table}[t]
    \centering
    \resizebox{1.0\columnwidth}{!}{
    \begin{tabular}{ll ccc}
        \toprule
        \textbf{Model} & \textbf{Method} & \textbf{Seen Tokens $\uparrow$} & \textbf{Unseen Tokens $\uparrow$} & \textbf{All Tokens $\uparrow$} \\
        \midrule
        \multirow{3}{*}{Llama3-3B}
        & Text & $93.6 \pm 0.2\%$ & $86.7 \pm 0.5\%$ & $92.0 \pm 0.2\%$ \\
        & Steering token & $94.5 \pm 0.2\%$ & $89.6 \pm 0.4\%$ & $93.4 \pm 0.2\%$ \\
        & Hybrid & $\mathbf{96.3 \pm 0.2\%}$ & $\mathbf{92.6 \pm 0.4\%}$ & $\mathbf{95.5 \pm 0.1\%}$ \\
        \midrule
        \multirow{3}{*}{Llama3-8B}
        & Text & $93.0 \pm 0.2\%$ & $81.2 \pm 0.5\%$ & $90.4 \pm 0.2\%$ \\
        & Steering token & $93.4 \pm 0.2\%$ & $83.2 \pm 0.5\%$ & $91.1 \pm 0.2\%$ \\
        & Hybrid & $\mathbf{95.7 \pm 0.2\%}$ & $\mathbf{88.7 \pm 0.5\%}$ & $\mathbf{94.1 \pm 0.2\%}$ \\
        \midrule
        \multirow{3}{*}{Qwen3-4B}
        & Text & $\mathbf{96.6 \pm 0.2\%}$ & $82.1 \pm 0.5\%$ & $93.2 \pm 0.2\%$ \\
        & Steering token & $96.1 \pm 0.2\%$ & $80.6 \pm 0.5\%$ & $92.7 \pm 0.2\%$ \\
        & Hybrid & $\mathbf{96.6 \pm 0.1\%}$ & $\mathbf{84.3 \pm 0.5\%}$ & $\mathbf{93.8 \pm 0.2\%}$ \\
        \midrule
        \multirow{3}{*}{Qwen3-8B}
        & Text & $\mathbf{94.8 \pm 0.2\%}$ & $81.6 \pm 0.5\%$ & $91.7 \pm 0.2\%$ \\
        & Steering token & $94.7 \pm 0.2\%$ & $\mathbf{83.2 \pm 0.5\%}$ & $\mathbf{92.0 \pm 0.2\%}$ \\
        & Hybrid & $94.1 \pm 0.2\%$ & $82.6 \pm 0.5\%$ & $91.5 \pm 0.2\%$ \\
        \bottomrule
    \end{tabular}
    }
    \caption{Single-behavior performance on seen, unseen, and all tokens across models.}
    \label{tab:single-token-performance}
\end{table}


\section{Interpretability of the Steering Tokens}\label{app:interpretability}
We have conducted two interpretability analyses to analyze the meaning of the steering tokens.

\textbf{Nearest vocabulary neighbors.} We computed the top-5 nearest vocabulary tokens by cosine similarity for each learned steering token (Qwen3-8B and Llama3-8B) and observed the following:
\textbf{Language tokens:} nearest neighbor is the corresponding language name (e.g., embedding of ``French'' is the closest to the embedding of the steering token \texttt{<french>} at ~0.44–0.52 cosine sim).
\textbf{Length tokens:} nearest neighbors are digit tokens matching the numeric boundaries (e.g., ``50'', ``70'').
\textbf{\texttt{<and> token}:} nearest neighbors have cosine similarity $\sim 0.09-0.10$ on both models---effectively near-random. The learned composition embedding does not align with any interpretable vocabulary region. It is placed in a new, distinct region of the model's embedding space.
This is expected, given our design: behavior tokens are initialized as the mean of the word embeddings of their corresponding instruction, so they retain proximity to semantically related vocabulary. The \texttt{<and>} token, by contrast, is initialized to the zero vector and explicitly trained to be orthogonal to all seen behavior tokens---so its substantial distance from interpretable vocabulary is expected, not incidental.

\begin{figure}[t]
  \centering
  \includegraphics[width=\linewidth]{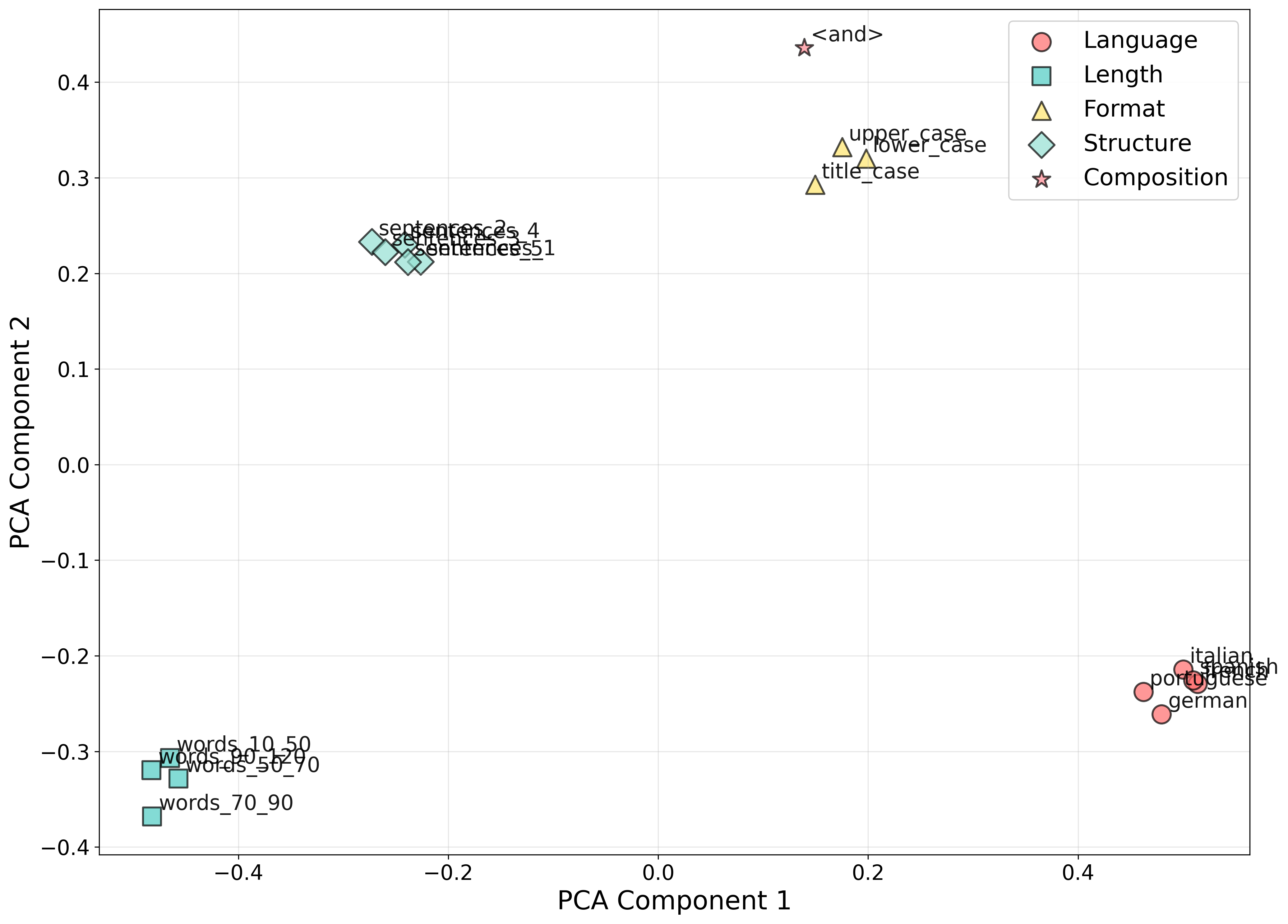}
  \caption{2D PCA projection of the Qwen3-14B steering token embeddings.}
  \label{fig:qwen14b_pca}
\end{figure}

\textbf{PCA clustering.} We project all learned steering tokens of Qwen3-14B into 2D, behavior tokens cluster tightly by category (language, length, format, structure). We observe that the \texttt{<and>} token is a clear outlier, isolated from all behavior token clusters. See Figure~\ref{fig:qwen14b_pca}.

Together, these results support the claim that \texttt{<and>} learns a structurally distinct function of composing (arbitrary) behavior, whereas the individual behavior tokens essentially compress the semantics of the corresponding instructions. \texttt{<and>} token's embedding is not a rephrasing of any single behavior or of the vocabulary token ``and.''

\section{LLM-judge for Response Quality Evaluation}\label{app:llm-as-judge}
\begin{prompt}
\textbf{System Prompt}

You are an expert evaluator of AI-generated responses.  
Your task is to assess the quality of a model's answer to a question, taking
into account the instruction that was provided when generating the answer.

Do not evaluate whether the instruction was followed—assume it was.
Instead, use the instruction to set realistic expectations for the scope,
depth, style, and language of the answer.

\medskip
\textbf{Examples}

\begin{itemize}
    \item If the instruction says to answer in 10--50 words, do not penalize brevity.
    \item If the instruction says to answer in a different language (e.g., Spanish),
    do not evaluate the answer based on English fluency.
\end{itemize}

\medskip
\textbf{Rating Scale}

\begin{itemize}
    \item 1 = Very poor
    \item 2 = Poor
    \item 3 = Acceptable
    \item 4 = Good
    \item 5 = Excellent
\end{itemize}

Output only the number. Do not include any explanation.

\medskip
\hrule
\medskip

\textbf{Evaluation Template}

\begin{description}
    \item[Instruction:] \texttt{\{instruction\}}
    \item[Question:] \texttt{\{question\}}
    \item[Model's Answer:] \texttt{\{answer\}}
    \item[Evaluation:]
\end{description}
\end{prompt}

\section{Metrics}\label{app:metrics}

\textbf{Mean and best accuracy}: For a $k$-property combination, accuracy measures the percentage of responses satisfying all $k$ constraints simultaneously (evaluated on $n=1000$ samples). Since token orderings can affect performance, we evaluate all $k!$ possible orderings and report the mean accuracy across orderings as our primary metric. We also report \textbf{best accuracy} (maximum across orderings) to characterize the upper bound achievable with optimal ordering.

\textbf{Order variance}: For each token combination tested with multiple orderings, we compute the maximum pairwise absolute difference in accuracy: $\Delta_{\text{max}} = \max_{i,j} |a_i - a_j|$, where $a_i$ and $a_j$ are accuracies for different orderings of the same token set. We report two statistics: (1) the mean of $\Delta_{\text{max}}$ across all token combinations with multiple orderings (Avg), and (2) the maximum $\Delta_{\text{max}}$ observed across all combinations (Max). Lower variance indicates more robust, order-invariant compositional behavior. This metric is critical for practical deployment scenarios, where users cannot manually optimize token ordering for each task. Unlike standard deviation, which measures spread around the mean, the maximum pairwise difference captures the worst-case sensitivity to ordering—directly quantifying the risk that an unfavorable token arrangement could substantially degrade performance. For a truly order-invariant composition mechanism, $\Delta_{\text{max}}$ should approach zero regardless of token arrangement.

\section{Baseline Implementation Details}\label{app:baselines}
We document the three parameter-based baselines compared against steering tokens in Table~\ref{tab:steering-vs-baselines}: \textbf{LoRA DARE}, \textbf{LM-Steer} output steering baseline, and \textbf{CAA}. We train LoRA DARE and LM-Steer using the same self-distillation objective, training data, and training schedule as steering tokens (see \S\ref{app:implementation-details}), so any difference in performance reflects the steering method rather than the supervision signal. The base LLM is frozen in both cases; only the baseline-specific parameters are updated. For the \textbf{Contrastive Action Addition} baseline \cite{rimsky2024steering} we use their official codebase: \url{https://github.com/nrimsky/CAA}

\subsection{LoRA DARE Baseline}\label{app:lora-dare}

\rparagraph{Per-behavior adapter} For each behavior $b$, we train a separate LoRA adapter~\citep{hu2022lora} with rank $r=8$ and scaling factor $\alpha=16$, dropout $0.1$, and no additive bias (\texttt{bias=``none''}). The adapter is inserted \emph{only into the MLP sublayer projections} of every transformer block---specifically the modules \texttt{gate\_proj}, \texttt{up\_proj}, and \texttt{down\_proj}----while attention projections and embedding/LM-head matrices remain unmodified and frozen.

\rparagraph{Composition via DARE Linear} For multi-behavior steering we merge the per-behavior adapters into a single adapter using \textbf{DARE Linear}~\citep{yu2024language}: adapter weights are randomly dropped with probability $p_{\text{drop}}$ and the surviving entries are rescaled by $1/(1-p_{\text{drop}})$, then combined by a linear (weighted-average) aggregation of the sparsified updates. The merged adapter is applied as a single PEFT module at inference. This is a merging procedure used to produce the 2- and 3-behavior numbers in Table~\ref{tab:steering-vs-baselines}. No separate adapter is trained per combination; composition is purely post-hoc.

\rparagraph{Inference} Inputs are prepared with the model's native chat template. Generation uses vLLM with the same greedy-decoding configuration as our main experiments (temperature $0$, up to $256$ tokens), with \texttt{max\_loras} set to accommodate all evaluated combinations.

\rparagraph{Order-variance reporting} Because DARE Linear merging is invariant to the order of behaviors, we leave the Ord.\ Var. column in Table~\ref{tab:steering-vs-baselines} blank for this baseline.

\subsection{LM-Steer (Output Steering) Baseline}\label{app:lm-steer}

\rparagraph{Intervention} Following \citet{han2024word}, the LM-Steer baseline inserts a learned linear transformation into the \emph{final hidden state} of the transformer, immediately before the LM head:
\[
\mathbf{h}' \;=\; \mathbf{h} + \varepsilon\, \mathbf{W}\, \mathbf{h}, \qquad \varepsilon = 10^{-3}.
\]
The weight matrix $\mathbf{W} \in \mathbb{R}^{d \times d}$ (with $d$ the model hidden size) is parameterised via a rank-$r$ factorisation
\[
\varepsilon \mathbf{W} \;\approx\; \mathbf{P}_1 \mathbf{P}_2, \qquad \mathbf{P}_1 \in \mathbb{R}^{d \times r},\; \mathbf{P}_2 \in \mathbb{R}^{r \times d},
\]
with $r=64$. Both factors are initialised as $\mathcal{N}(0, \sigma^2)$ with $\sigma = 0.01$ (the paper explicitly warns against zero-initialisation, which we respect). Only $\mathbf{P}_1$ and $\mathbf{P}_2$ are trainable; all other model parameters are frozen. Concretely, the forward pass runs the transformer backbone in the usual way, takes the final \texttt{last\_hidden\_state}, applies $\mathbf{h}' = \mathbf{h} + \mathbf{h}\,\mathbf{P}_1\,\mathbf{P}_2$, and then feeds $\mathbf{h}'$ into the original (frozen) \texttt{lm\_head} to obtain logits. The loss is the standard causal-LM loss on $\mathbf{h}'$-derived logits, which backpropagates only into $\mathbf{P}_1$ and $\mathbf{P}_2$.

\rparagraph{Per-behavior matrices} We train one pair $(\mathbf{P}_1^{(p)}, \mathbf{P}_2^{(p)})$ per behavior $b$, using the same 50k-sample self-distillation data, optimiser, and schedule as our steering tokens. The trained matrices cover all 15 properties reported in the main experiments (\textit{languages}, \textit{format}, \textit{length}, \textit{structure}).

\rparagraph{Additive composition} For multi-behavior inference, the loaded per-behavior factor pairs are \emph{added} in parameter space before application---matching the additive composition recipe of \citet{han2024word}:
\[
\mathbf{h}' \;=\; \mathbf{h} + \mathbf{h}\, \Bigl(\sum_{p \in S} \mathbf{P}_1^{(p)}\Bigr) \Bigl(\sum_{p \in S} \mathbf{P}_2^{(p)}\Bigr),
\]
where $S$ is the set of active behaviors.


\rparagraph{Inference} As for LoRA, the prompt contains no steering tokens. Because LM-Steer has no ordering at the prompt level and sum-of-factors composition is symmetric (commutative under addition), there is again no ordering axis to average over, so the Ord. Var. column is left blank in Table~\ref{tab:steering-vs-baselines}. Inference uses greedy decoding with up to $256$ tokens

\subsection{CAA Baseline}\label{app:caa}

\rparagraph{Steering vector construction} Following \citet{rimsky2024steering}, we construct one steering vector per behavior from hidden-state differences. For each behavior $b$, we collect paired positive and negative examples $D_b$, extract the hidden state at the last token position from layer $l$, compute the activation difference for each pair, and define the behavior steering vector as the mean of these differences:
\[
\mathbf{v}^{(b)} \;=\; \frac{1}{|D_b|} \sum_{(x^{+}, x^{-}) \in D_b}
\Bigl(
\mathbf{h}^{(l)}_{\mathrm{last}}(x^{+})
-
\mathbf{h}^{(l)}_{\mathrm{last}}(x^{-})
\Bigr)
\]

\rparagraph{Composition} For multi-behavior steering, we compose the per-behavior vectors additively. Given a set of active behaviors $S$, we form a single steering vector as a weighted sum
\[
\mathbf{v}_{S} \;=\; \sum_{b \in S} \alpha_b \mathbf{v}^{(b)},
\]
where in our experiments all active behaviors use the same multiplier $\alpha_b = 3.0$.

\rparagraph{Inference} During generation, we add the composed steering vector to the model hidden state at the same transformer layer used for extraction: $\mathbf{h}^{(l)}\leftarrow \mathbf{h}^{(l)} + \mathbf{v}_{S}$. Thus, unlike LoRA DARE and LM-Steer, CAA performs activation-space intervention directly at inference time rather than learning additional trainable parameters.

\rparagraph{Hyperparameters} We extract and apply steering vectors at layer $l=10$. We set the steering multiplier to $3.0$, selected based on single-behavior performance on a small validation set. Generation uses greedy decoding with temperature $0$ and a maximum of $256$ new tokens.

\rparagraph{Order-variance reporting} Because CAA composes behaviors by summing steering vectors, the composition rule is invariant to behavior order. We therefore leave the Ord.\ Var. column blank for this baseline in Table~\ref{tab:steering-vs-baselines}.

\section{Instruction Template Variants}\label{app:instruction-variants}

\begin{tcolorbox}[appendixbox, title=Language Instruction Variants (\{language\})]
\small
\setlength\itemsep{1pt}
\begin{itemize}
  \item Answer in \{language\}.
  \item Respond in \{language\}.
  \item Provide your answer in \{language\}.
  \item Reply in \{language\}.
  \item Give your response in \{language\}.
  \item Write your answer in \{language\}.
  \item Use \{language\} for your response.
  \item Provide a \{language\} response.
  \item Respond using \{language\}.
  \item Answer using the \{language\} language.
\end{itemize}
\end{tcolorbox}

\begin{tcolorbox}[
  appendixbox,
  title=\detokenize{Word Range Instruction Variants ({min}, {max})}
]
\small
\setlength\itemsep{1pt}
\begin{itemize}
  \item Answer in \{min\} to \{max\} words.
  \item Respond in \{min\} to \{max\} words.
  \item Keep your response between \{min\} and \{max\} words.
  \item Use \{min\}-\{max\} words in your answer.
  \item Provide an answer of \{min\} to \{max\} words.
  \item Your response should contain \{min\} to \{max\} words.
  \item Limit your answer to \{min\}-\{max\} words.
  \item Give a response that is \{min\} to \{max\} words long.
  \item Answer using between \{min\} and \{max\} words.
  \item Provide a \{min\} to \{max\} word response.
\end{itemize}
\end{tcolorbox}

\begin{tcolorbox}[appendixbox, title=Sentence Count Instruction Variants (\{num\})]
\small
\setlength\itemsep{1pt}
\begin{itemize}
  \item Provide a response in \{num\} sentences.
  \item Answer in \{num\} sentences.
  \item Give a \{num\}-sentence response.
  \item Respond using exactly \{num\} sentences.
  \item Your answer should be exactly \{num\} sentences.
  \item Limit your response to \{num\} sentences.
  \item Provide an answer consisting of \{num\} sentences.
  \item Reply with \{num\} sentences.
  \item Use \{num\} sentences for your answer.
  \item Answer in \{num\} sentences only.
\end{itemize}
\end{tcolorbox}

\begin{tcolorbox}[appendixbox, title=Case Formatting Instruction Variants (\{case\_type\})]
\small
\setlength\itemsep{1pt}
\begin{itemize}
  \item Each word in the answer should be in \{case\_type\}.
  \item Write your answer with each word in \{case\_type\}.
  \item Use \{case\_type\} for every word in your response.
  \item Use only \{case\_type\} letters in your answer.
  \item Format your answer in \{case\_type\}.
  \item All words should be in \{case\_type\}.
  \item Apply \{case\_type\} formatting to your answer.
  \item Your response should use \{case\_type\} for all words.
  \item Make all letters \{case\_type\}.
  \item Use \{case\_type\} throughout your answer.
\end{itemize}
\end{tcolorbox}

\begin{tcolorbox}[appendixbox, title=Json Wrap Instruction Variants (\{json\_format\})]
\small
\setlength\itemsep{1pt}
\begin{itemize}
  \item Wrap your answer in JSON with the key 'answer'.
  \item Return your response as a JSON object with an 'answer' field.
  \item Format your reply as JSON: \{\"answer\": \"...\"\}.
  \item Respond with a JSON object containing an 'answer' key.
  \item Output your answer as JSON with the key 'answer'.
  \item Provide your response in JSON format using the key 'answer'.
  \item Give your answer inside a JSON object with an 'answer' field.
  \item Structure your response as \{\"answer\": \"your response\"\}.
  \item Reply in JSON format with your answer under the 'answer' key.
  \item Encode your response as a JSON object with an 'answer' property.
\end{itemize}
\end{tcolorbox}

\section{Model Generations}\label{app:generations}
Table~\ref{tab:example_generations} presents 19 representative examples of \textbf{compositional control} using steering tokens across three model sizes (Qwen-4B, Qwen-8B, Qwen-14B). All examples shown are success cases where all specified constraints are satisfied simultaneously.

\begin{table*}[t]
\centering
\tiny
\setlength{\tabcolsep}{3pt}
\begin{tabular}{l l p{0.30\linewidth} p{0.38\linewidth}}
\toprule
\textbf{Model} & \textbf{Steering Tokens} & \textbf{Prompt} & \textbf{Generation} \\
\midrule
Qwen-8B & \texttt{spanish} \texttt{words\_10\_50} & What are the most effective ways to reduce energy consumption in a small office setting as part of an environmental... & Para reducir el consumo energético en una oficina pequeña, utiliza luces LED, apaga dispositivos no en uso, usa computadoras en modo de suspensión, y realiza mantenimiento regular... \\
Qwen-8B & \texttt{french} \texttt{words\_10\_50} & In what ways does the Mediterranean diet compare to the American diet in terms of reducing the risk of heart attacks? & La diète méditerranéenne réduit davantage le risque d'infarctus en privilégiant les fruits, légumes, oléagineux et poissons, tandis que la diète américaine, riche en graisses... \\
Qwen-8B & \texttt{lower\_case} \texttt{words\_10\_50} & How should I structure a follow-up email after accepting a job offer to confirm the details? & thank you for the offer. i confirm my acceptance and appreciate the opportunity. please confirm the start date, role, and any additional details. looking forward to joining the... \\
Qwen-8B & \texttt{upper\_case} \texttt{french} & What are the advantages of using cloud-based [application] over on-premise solutions for [company]? & LES AVANTAGES DE L'UTILISATION DE SOLUTIONS EN CLOUD POUR [COMPANY] PAR RAPPORT AUX SOLUTIONS HÉBERGÉES EN INTERNE SONT LES SUIVANTS : PLUS GRANDE FLEXIBILITÉ, COÛTS... \\
Qwen-8B & \texttt{words\_10\_50} \texttt{portuguese} & What does the theme of love versus infatuation reveal about Romeo and Juliet's relationship in the play? & O tema do amor versus a infatuação revela que o relacionamento de Romeo e Julieta é baseado em paixão intensa, mas imatura, que se transforma em tragédia devido à falta de...  \\
Qwen-8B & \texttt{upper\_case} \texttt{words\_10\_50} & In what ways does wind power contribute to lowering greenhouse gas emissions compared to traditional fossil fuel-based... & WIND POWER GENERATES CLEAN ENERGY WITHOUT BURNING FUELS, REDUCING COAL, OIL, AND GAS USE. IT CUTS CARBON DIOXIDE AND OTHER GREENHOUSE GASES, HELPING MITIGATE CLIMATE CHANGE. NO... \\
Qwen-8B & \texttt{lower\_case} \texttt{spanish} \texttt{sentences\_4} & What are the key steps to ensure a successful nonprofit organization launch focused on environmental conservation? & primero, define un propósito claro y una misión específica relacionada con la conservación ambiental. segundo, desarrolla un plan de acción detallado que incluya estrategias de... \\
Qwen-8B & \texttt{sentences\_5} \texttt{lower\_case} \texttt{spanish} & Can you provide historical examples where perseverance was key to achieving major breakthroughs? & en la historia, la perseverancia ha sido clave para lograr avances importantes. maria curie, a pesar de las dificultades y la falta de apoyo, logró descubrir el radio y el... \\
\midrule
Qwen-4B & \texttt{portuguese} \texttt{words\_10\_50} & What are the main advantages of using a project management tool for team collaboration? & Ferramentas de gerenciamento de projetos facilitam a comunicação, centralizam tarefas, melhoram a transparência e aumentam a produtividade, permitindo que equipes trabalhem de... \\
\midrule
Qwen-14B & \texttt{german} \texttt{words\_10\_50} & Should a first-aid kit be considered one of the top 5 essential items for a beach vacation, and what should it include? & Ja, ein Erste-Hilf-Kasten gehört zu den Top 5 für einen Badeurlaub. Er sollte Verbände, Desinfektionsmittel, Schmerztabletten, Pflaster und eine Pinzette enthalten, um... \\
Qwen-14B & \texttt{italian} \texttt{words\_10\_50} & How can I effectively practice behavioral interview questions for a role in the engineering industry? & Per prepararti efficacemente alle domande comportamentali per un ruolo ingegneristico, ripassa esperienze passate, utilizza il metodo STAR (Situazione, Compito, Azione, Risultato)... \\
Qwen-14B & \texttt{spanish} \texttt{lower\_case} & How do bullet points contribute to the accessibility of written content for individuals with reading difficulties? & los puntos de viñeta ayudan a hacer más accesible el contenido escrito para personas con dificultades de lectura al dividir la información en partes más pequeñas y manejables.... \\
Qwen-14B & \texttt{words\_10\_50} \texttt{italian} & How should I structure a follow-up email after accepting a job offer to confirm the details? & Gentile [Nome], confermo l'accettazione dell'offerta di lavoro. Posso confermare i dettagli dell'incarico, della retribuzione e della data di inizio? Resto a disposizione per... \\
Qwen-14B & \texttt{french} \texttt{words\_10\_50} \texttt{lower\_case} & What strategies can a manager use to enhance workplace efficiency, boost employee productivity, and foster better... & utiliser des outils de gestion du temps, encourager la communication ouverte, organiser des formations régulières, promouvoir l'autonomie et reconnaître les performances pour... \\
Qwen-14B & \texttt{words\_10\_50} \texttt{lower\_case} \texttt{spanish} & In what ways can a project management tool help a school administrator organize and track the progress of various... & una herramienta de gestión de proyectos puede ayudar a un administrador escolar a organizar tareas, asignar responsables, establecer plazos y monitorear el progreso de proyectos... \\
Qwen-14B & \texttt{sentences\_1} \texttt{lower\_case} \texttt{italian} & I'm planning a trip to Bangkok. What are the best places to experience local culture, and what should I be cautious... & i migliori posti per esperire la cultura locale a bangkok sono il mercato fluttuante di damnoen saduak, il tempio di wat pho e il quartiere di chinatown, mentre si deve essere... \\
Qwen-14B & \texttt{sentences\_1} \texttt{lower\_case} \texttt{french} & Are there specific types of exercise that are most effective for improving mental health in teenagers? & les exercices aérobiques réguliers, comme la course à pied, la natation ou le vélo, sont particulièrement efficaces pour améliorer la santé mentale des adolescents. \\
Qwen-14B & \texttt{lower\_case} \texttt{italian} \texttt{words\_10\_50} & What role does Mr. Darcy play in the development of the main plot in *Pride and Prejudice*? & mr. darcy è centrale nel plot principale di "orgoglio e pregiudizio". il suo rapporto con elizabeth evoluisce da ostilità a amore, influenzando gli eventi chiave, come il... \\
Qwen-14B & \texttt{lower\_case} \texttt{sentences\_1} \texttt{italian} & What is the capital of France, and what is the name of the famous department store that is both a shopping destination... & la capitale della francia è parigi e il famoso centro commerciale che è anche un attrazione turistica si chiama le bon marché.\\
\bottomrule
\end{tabular}
\caption{Example generations demonstrating compositional control with steering tokens.}
\label{tab:example_generations}
\end{table*}

\end{document}